\documentclass[10pt,twocolumn,letterpaper]{article}

\usepackage[pagenumbers]{cvpr} 

\definecolor{cvprblue}{rgb}{0.21,0.49,0.74}
\usepackage[pagebackref,breaklinks,colorlinks,allcolors=cvprblue]{hyperref}
\usepackage{makecell}
\usepackage{multicol}
\usepackage{multirow}
\usepackage{algorithm}
\usepackage{makecell}
\usepackage{algpseudocode}
\algrenewcommand\algorithmicrequire{\textbf{Input:}}
\algrenewcommand\algorithmicensure{\textbf{Output:}}
\def\paperID{} 
\def\confName{CVPR}
\def\confYear{2026}

\title{Spatia: Video Generation with Updatable Spatial Memory\vspace{-3mm}} 

\author{
    Jinjing Zhao\textsuperscript{1}\textsuperscript{*} \quad
    Fangyun Wei\textsuperscript{2}\textsuperscript{*}\textsuperscript{$\dagger$} \quad
    Zhening Liu\textsuperscript{3} \quad
    Hongyang Zhang\textsuperscript{4} \quad
    Chang Xu\textsuperscript{1}\textsuperscript{$\dagger$} \quad
    Yan Lu\textsuperscript{2}
    \protect\vspace{1mm} \protect\\ 
    \textsuperscript{1}The University of Sydney \qquad
    \textsuperscript{2}Microsoft Research \qquad
    \textsuperscript{3}HKUST \qquad
    \textsuperscript{4}University of Waterloo \\
    {\small \url{https://zhaojingjing713.github.io/Spatia/}}
}

\begin{document}
\twocolumn[{%
\renewcommand\twocolumn[1][]{#1}%
\maketitle
\vspace{-10mm}
\begin{center}
    \centering
    \captionsetup{type=figure}
    \includegraphics[width=\textwidth]{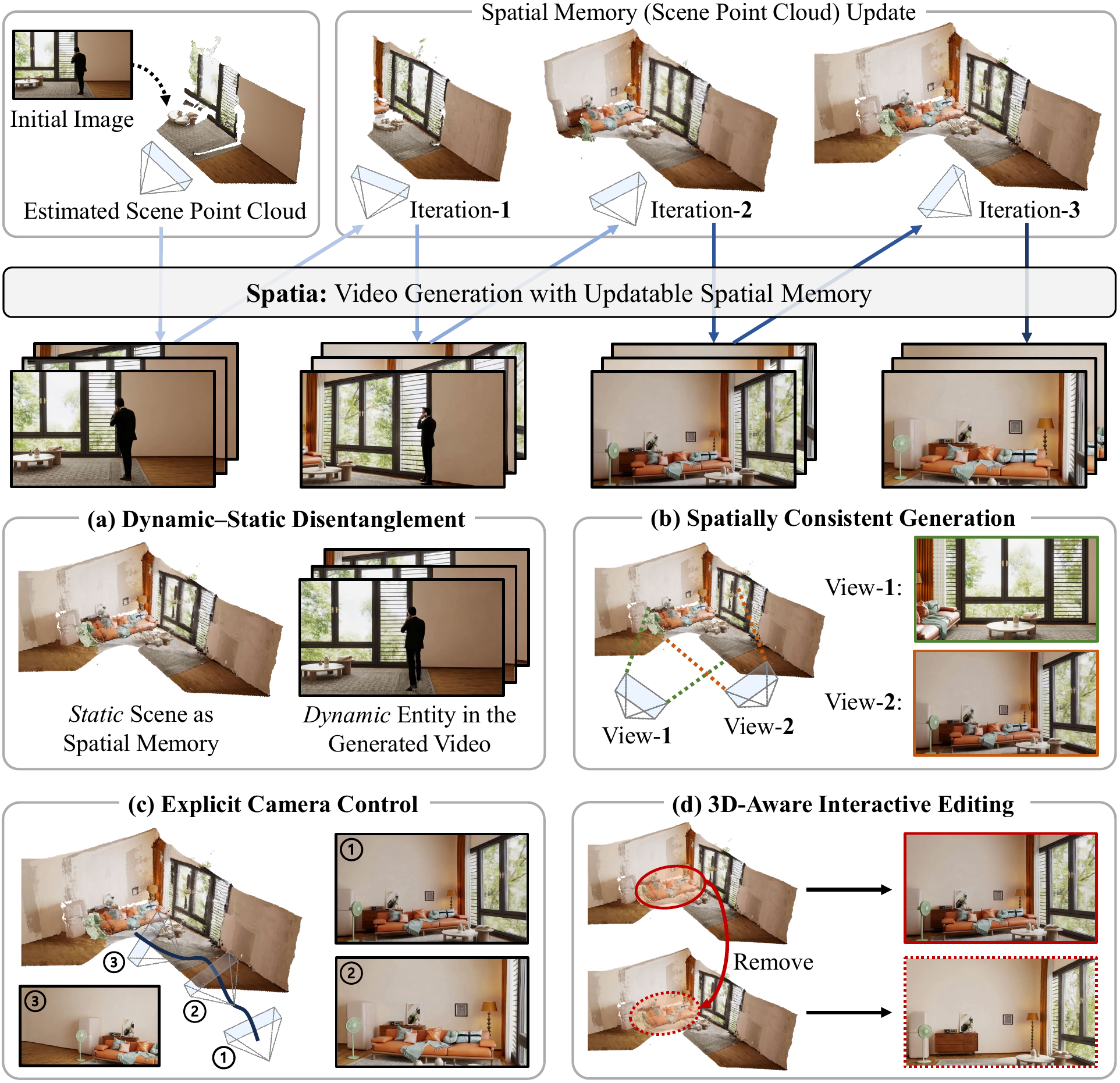}
    \vspace{-7mm}
    \captionof{figure}{\textbf{Spatia} maintains a scene point cloud as its spatial memory and conditions on it throughout the iterative video generation process. It enables: (a) dynamic–static disentanglement by treating the static scene as spatial memory while generating videos that include dynamic entities; (b) spatially consistent generation across multiple views; (c) explicit camera control via 3D-aware trajectory rendering; and (d) 3D-aware interactive editing, allowing users to modify or remove scene elements prior to video generation.}
    \label{fig:teaser}
\end{center}%

\noindent
\begin{minipage}{\textwidth}
    \noindent\rule{3.4cm}{0.4pt}\\ 
    \footnotesize \textsuperscript{*}Equal contribution.\\
    \footnotesize \textsuperscript{$\dagger$}Corresponding author.
\end{minipage}
}]

\begin{abstract}
Existing video generation models struggle to maintain long-term spatial and temporal consistency due to the dense, high-dimensional nature of video signals. To overcome this limitation, we propose Spatia, a spatial memory–aware video generation framework that explicitly preserves a 3D scene point cloud as persistent spatial memory. Spatia iteratively generates video clips conditioned on this spatial memory and continuously updates it through visual SLAM. This dynamic–static disentanglement design enhances spatial consistency throughout the generation process while preserving the model’s ability to produce realistic dynamic entities. Furthermore, Spatia enables applications such as explicit camera control and 3D-aware interactive editing, providing a geometrically grounded framework for scalable, memory-driven video generation.
\end{abstract}    
\vspace{-4.5mm}
\section{Introduction}
\label{sec:intro}
\vspace{-0.5mm}
Video generation has emerged as a foundational technique powering a wide spectrum of tasks. On one hand, recent advances in video generation foundation models~\cite{yang2024cogvideox,kong2024hunyuanvideo,wan2025wan,team2025longcat,sora,veo,gen3,Hailuo,hacohen2024ltx,gao2025seedance,Kling} have significantly improved the quality and controllability of short-duration video synthesis. On the other hand, there is a growing need to extend these models toward long-horizon video generation, enabling applications that require \textit{temporal consistency} and \textit{persistent memory}, such as world models~\cite{alonso2024diffusion, valevski2024diffusion, xiao2025worldmem, parkerholder2024genie2, yu2024wonderjourney, chen2025flexworld,zhang2025mega,engstler2024invisible,yu2024wonderworld}, AI-driven game generation~\cite{che2024gamegen, yang2024playable, oasis2024, bruce2024genie, feng2024matrix, yu2025gamefactory,zhang2025matrixgame,sun2025virtual}, and embodied AI~\cite{
Brohan2023RT2VM,Li2023VisionLanguageFM,Kim2024OpenVLAAO,Hancock2024RuntimeOI,Zhen20243DVLAA3,Li2025PointVLAIT,Team2024OctoAO,Wen2024DiffusionVLAGA,shenvideovla}.

Unlike LLMs~\cite{achiam2023gpt,openai2025gpt41,meta2025llama4,bai2023qwen,yang2025qwen3,liu2024deepseek,team2025kimi}, video generation models encounter intrinsic difficulties in encoding long-term historical information, primarily due to the dense and high-dimensional nature of video signals. For instance, a short 5-second 480P ($640 \times 480$) video at 24 FPS—consisting of 120 frames—already corresponds to $40\times 30 \times 30 = 36,000$ spatio-temporal tokens when using a video encoder~\cite{wan2025wan,kong2024hunyuanvideo} with a spatial downsampling factor of 16 and a temporal downsampling factor of 4. Including even one additional 5-second video clip as context would dramatically increase the computational and memory demands, rendering it impractical to directly model minute- or hour-scale temporal contexts, which could easily span millions of tokens.

By comparison, $36,000$ tokens can represent around $27,000$ words\footnote{A word is represented by an average of 1.3 tokens using GPT-3's text tokenizer.}. In other words, with the same number of tokens, a video generation model can capture only about 5 seconds of visual history, whereas an LLM can encompass a context equivalent to $27,000$ words. Therefore, unlike LLMs that can directly attend to all historical text tokens, video generation models must rely on alternative mechanisms to preserve memory and encode contextual dependencies without simply modeling the entire sequence of historical spatio-temporal tokens.

In this work, we introduce an explicit memory mechanism designed to achieve consistent and long-horizon video generation, particularly in scenarios where the same location reappears multiple times during the generation process. As illustrated in Figure~\ref{fig:teaser}, we take the image-to-video task as an illustrative example. The process begins by estimating an initial 3D scene point cloud from the conditional input image, which serves as the spatial memory of the scene. We then iteratively perform two key steps:
\begin{enumerate}
    \item Generate a new video clip conditioned on both the current 3D scene point cloud and the previously generated video clip, ensuring temporal and spatial consistency across iterations.
    \item Update the scene point cloud using visual SLAM algorithms based on both newly generated and previously generated frames, thereby incorporating new content while preserving existing scene information.
\end{enumerate}
This iterative update enables the system to maintain scene consistency and geometric coherence over long sequences, allowing the model to effectively ``remember'' previously visited locations. As a result, it can generate videos with realistic long-term structural continuity, yielding a persistent spatial memory of the scene. We name our approach as \textbf{Spatia}, short for spatial memory–aware video generation. Spatia enjoys the following key characteristics, which arise from the integration of the spatial memory mechanism:
\begin{itemize}
    \item \textit{Dynamic–Static Disentanglement} (Figure~\ref{fig:teaser}(a)). Spatia preserves a scene point cloud as spatial memory while simultaneously generating dynamic entities that interact coherently with the scene. This contrasts with previous methods~\cite{huang2025voyager, li2025vmem, yu2024viewcrafter, yu2025trajectorycrafter} addressing the video generation memory problem, which are typically limited to producing videos with static scenes only.
    \item \textit{Spatially Consistent Generation} (Figure~\ref{fig:teaser}(b)). By retrieving spatial memory, Spatia can generate diverse video sequences depicting the same location from different viewpoints while preserving a consistent spatial structure.
    \item \textit{Explicit Camera Control} (Figure~\ref{fig:teaser}(c)). Unlike previous approaches~\cite{bai2025recammaster,guo2023animatediff,feng2024i2vcontrol,he2024cameractrl, he2025cameractrl,li2025vmem, yang2024direct} that encode camera trajectories into latent features and inject them into video generation models—an indirect strategy that may result in inaccurate or unstable control—Spatia achieves camera control in a more explicit and geometrically grounded manner. Mirroring the rendering process of 3DGS~\cite{kerbl20233d}, it directly applies the desired camera path to the 3D scene point cloud and renders a corresponding 2D point cloud sequence, which serves as a conditioning signal to guide video generation along the specified camera trajectory.
    \item \textit{3D-Aware Interactive Editing} (Figure~\ref{fig:teaser}(d)). Since Spatia conditions video generation on the 3D scene point cloud, users can interactively edit the scene before generation—for example, by removing or modifying specific objects. Such edits are directly reflected in the generated videos, enabling intuitive and fine-grained control over scene composition and content.
\end{itemize}

We experimentally demonstrate that Spatia, equipped with the proposed spatial memory mechanism, significantly enhances spatial consistency throughout the generation process, without compromising its ability to produce dynamic entities or the visual quality of generated videos. Additional benefits include enabling long-horizon generation and supporting applications such as spatial editing.

\vspace{-1mm}
\section{Related Works}
\label{sec:related-works}
\vspace{-1mm}
\noindent\textbf{Video Generation Models.}
The field of video generation has evolved rapidly, progressing from early UNet-based latent diffusion models~\cite{blattmann2023stable, chen2023videocrafter1, chen2024videocrafter2, zeng2024make} to large-scale Diffusion Transformers~\cite{peebles2023scalable, esser2024scaling}. This architectural transition has given rise to a new generation of powerful foundation models, including open-source systems~\cite{yang2024cogvideox, ma2025step, kong2024hunyuanvideo, wan2025wan, team2025longcat} and high-performance proprietary counterparts~\cite{sora, veo, Kling}.
While bidirectional models employing global spatio-temporal attention achieve impressive fidelity~\cite{kong2024hunyuanvideo, zheng2024open1, lin2024open_plan, peng2025open2, ma2025step, wan2025wan, Hailuo, gen3}, their quadratic computational complexity fundamentally limits them to short-clip generation. To generate arbitrarily long sequences, autoregressive frameworks~\cite{kim2024fifo, jin2024pyramidal, alonso2024diffusion, valevski2024diffusion, gao2024vid, henschel2024streamingt2v, wu2022nuwa} have been proposed, which iteratively synthesize new content conditioned on previously generated frames. Subsequent studies~\cite{chen2024diffusion, song2025history, yin2024slow, gu2025long, xie2024progressive, magi1, chen2025skyreels, liu2025infinitystar} further address the issue of error accumulation in long-horizon generation. While these methods achieve strong temporal coherence, they still lack an explicit spatial memory mechanism.

\noindent\textbf{Camera Control in Video Generation.}
Precise camera control has become a key goal in video synthesis. One line of work conditions generation on explicit camera parameters—for example, AnimateDiff~\cite{guo2023animatediff} employs motion LoRAs to learn specific camera trajectories. Other methods incorporate various camera representations, such as point trajectories or Plücker embeddings~\cite{sitzmann2021lfns}, directly into the generator~\cite{yang2024direct, feng2024i2vcontrol, he2024cameractrl, zhou2025stable, li2025vmem, he2025cameractrl}. For finer-grained control, geometry-aware approaches use 3D information—such as rendered point clouds—to provide dense spatial guidance for camera path generation~\cite{yu2024viewcrafter, yu2025trajectorycrafter, gu2025das, ren2025gen3c, yang2025omnicam}. Meanwhile, video editing–based methods achieve controllability by re-targeting existing footage to new viewpoints~\cite{bai2025recammaster} or by transferring camera motion from reference videos~\cite{luo2025camclonemaster}.

\noindent\textbf{Long-term Memory Modeling.}
A central strategy for improving the long-term memory capacity of LLMs lies in expanding their native context window, which has grown dramatically—from the limited spans of early models~\cite{radford2018improving, radford2019language, brown2020language, devlin2019bert, raffel2020exploring} to the million- or even ten-million-token ranges achieved by modern architectures~\cite{anthropic2024claude3, achiam2023gpt, team2024gemini, google2025gemini2flash, anthropic2025claude4, openai2025gpt41, meta2025llama4}—enabled by techniques like KV-cache compression \cite{xiong2023effective, ding2024longrope, kwon2023efficient, zhang2023h2o, li2024snapkv, liu2023ring}. In video generation, the bidirectional spatiotemporal attention used in most diffusion models prevents standard KV caching, thereby severely limiting the context window and restricting access to previously generated content. To preserve long-term spatial consistency, recent works have introduced memory-based architectures. For maintaining spatial coherence in explorable 3D scenes, methods such as~\cite{yu2024viewcrafter,chen2025flexworld, ma2025you,liu2024dynamics,huang2025voyager} leverage progressive expansion or warping pipelines to refine global scene geometry. Context-as-Memory~\cite{yu2025context} retrieves previous frames based on camera FOV overlap, while VMem~\cite{li2025vmem} introduces a surfel-indexed view memory for efficient geometric indexing and retrieval of past views. 

\noindent\textbf{Scene Point Cloud Estimation.}
Recent progress in visual geometry estimation is led by Dust3R~\cite{wang2024dust3r}, which unifies pairwise pose and geometry estimation but encounters a costly $\mathcal{O}(N^2)$ global alignment bottleneck when reconstructing an $N$-view input sequence. This limitation motivates follow-up works~\cite{mast3r_eccv24, cabon2025must3r, Yang_2025_Fast3R, wang2025continuous} to develop more scalable solutions by introducing efficient sequential or parallel architectures. In parallel, universal end-to-end models~\cite{wang2024vggsfm, wang2025vggt, wang2025pi, keetha2025mapanything} eliminate the pairwise dependency, employing large Transformers to infer globally consistent 3D geometry and camera parameters for all views in a single forward pass.
\vspace{-1mm}
\section{Method}
\label{sec:method}
\vspace{-1mm}
\begin{figure*}[!t]
    \centering
    \includegraphics[width=0.99\linewidth]{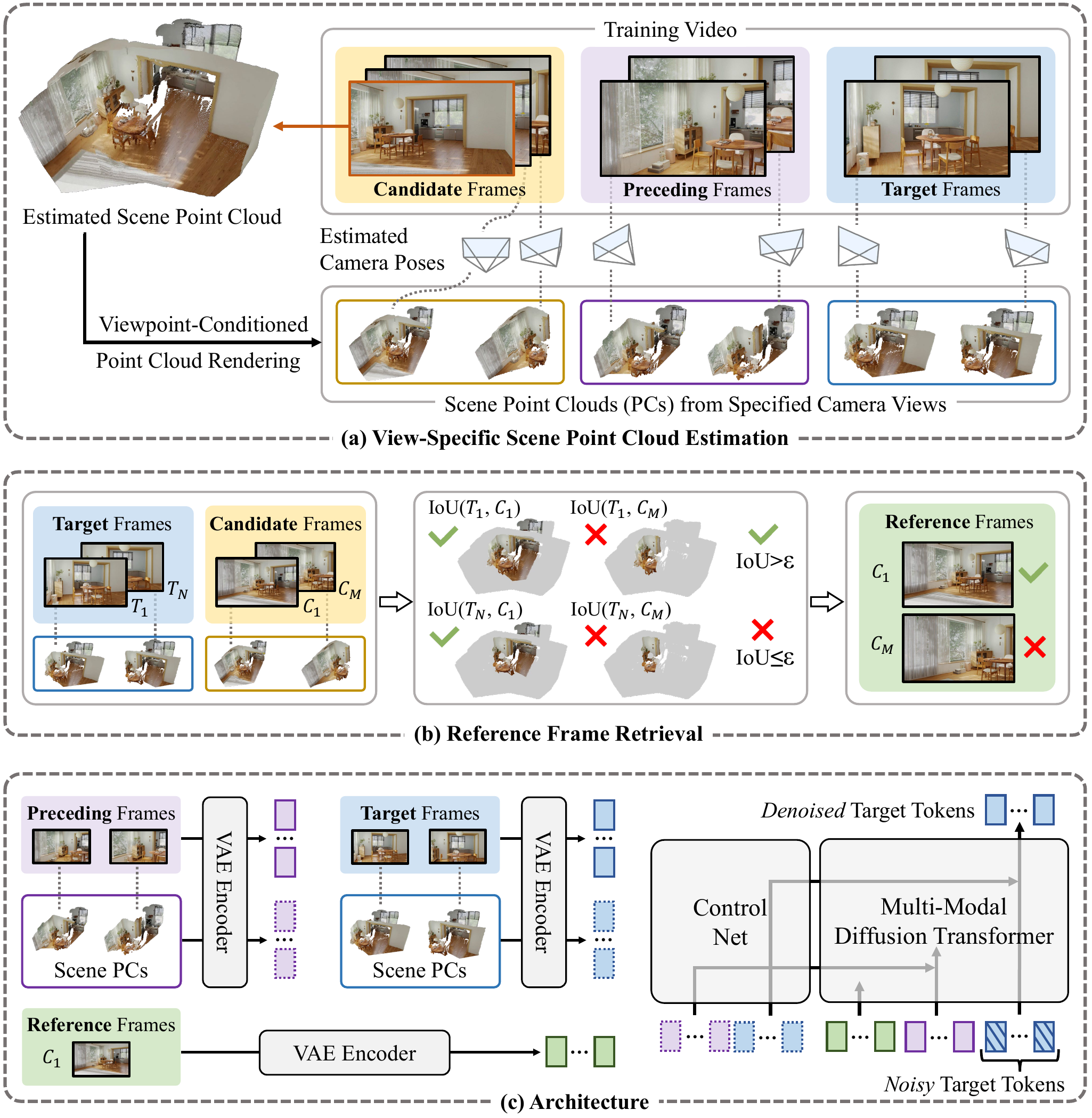}
    \vspace{-3.5mm}
    \caption{Overview of the training stage of Spatia. Each training video is divided into a target clip, a preceding clip, and a candidate-frame set. Text tokens are omitted for simplicity. (a) A frame is randomly selected from the candidate-frame set to estimate a 3D scene point cloud $\mathcal{S}$. Using the estimated camera poses together with $\mathcal{S}$, we then generate view-specific scene point cloud sequences for both the target and preceding clips. (b) The most spatially relevant frames are then retrieved from the candidate-frame set as reference frames. (c) The spatial conditions obtained from (a) and (b) guide the video generation process. The detailed network architecture is provided in Figure~\ref{fig:network}.}
    \vspace{-3.5mm}
    \label{fig:overview}
\end{figure*}

\noindent\textbf{Problem Formulation.} The objective of Spatia is to endow a video generation model with persistent spatial memory, enabling it to produce videos that are both spatially and temporally consistent. To achieve this, Spatia maintains and iteratively updates a static scene point cloud throughout the generation process. This point cloud serves as an explicit geometric memory that anchors all generated content within a coherent spatial layout. Spatia formulates the entire framework as a multi-modal conditional generation problem, where generation is conditioned on textual instructions, spatial memory, and temporal context. Specifically, the framework operates in two stages: 
\begin{enumerate}
    \item Generating a video clip conditioned on multi-modal inputs—including text instructions (for instruction following), geographically retrieved information from the spatial memory (for spatial consistency), and either an initial image or previously generated clips (for temporal continuity).
    \item Updating the spatial memory to incorporate newly generated content, ensuring that subsequent generations remain geometrically consistent with the evolving scene.
\end{enumerate}
Note that the above stages can be performed iteratively to enable long-horizon generation. Sections~\ref{sec:training} and~\ref{sec:inference} detail the training and inference of Spatia, respectively.

\vspace{-1mm}
\subsection{Training}
\label{sec:training}
\vspace{-1mm}
\noindent\textbf{Training Data.} We address the text-and-image-to-video generation problem, where each training sample consists of a video $\mathcal{V}$ paired with a textual description $\mathcal{T}$ that narrates its content. For a given training video $\mathcal{V}$, we decompose it into three parts, $\mathcal{V}= \{\boldsymbol{T}\}^{N} \cup \{\boldsymbol{P}\}^{M} \cup \{\boldsymbol{C}\}^{O}$, where $\{\boldsymbol{T}\}^{N}$, $\{\boldsymbol{P}\}^{M}$, and $\{\boldsymbol{C}\}^{O}$ denote the target-frame set, preceding-frame set, and candidate-frame set, containing $N$, $M$ and $O$ frames, respectively. Specifically, we randomly select one clip from $\mathcal{V}$ as the target clip $\{\boldsymbol{T}\}^{N}$, in which each $\boldsymbol{T}$ represents a target frame to be generated by the model. The clip immediately preceding $\{\boldsymbol{T}\}^{N}$ is defined as the preceding clip $\{\boldsymbol{P}\}^{M}$, with $\boldsymbol{P}$ referring to a single frame providing temporal context. The remaining frames within $\mathcal{V}$, excluding those in $\{\boldsymbol{T}\}^{N}$ and $\{\boldsymbol{P}\}^{M}$, are treated as candidate frames $\{\boldsymbol{C}\}^{O}$, which serve as potential references for spatial and geometric consistency.

\noindent\textbf{Overview.} Figure~\ref{fig:overview} illustrates the overall training pipeline, which can be divided into two main parts: data pre-processing—including \textit{View-Specific Scene Point Cloud Estimation} (Section~\ref{sec:PCE}) and \textit{Reference Frame Selection} (Section~\ref{sec:RFS})—and the model \textit{Architecture} (Section~\ref{sec:archi}), formulated as a multi-modal conditional generation framework.

\vspace{-1mm}
\subsubsection{View-Specific Scene Point Cloud Estimation}
\label{sec:PCE}
\textbf{Scene Point Cloud Estimation.} As shown in Figure~\ref{fig:overview}(a), we first randomly sample a frame from the candidate-frame set $\{\boldsymbol{C}\}^{O}$ and employ MapAnything~\cite{keetha2025mapanything} to estimate a scene point cloud $\mathcal{S}$. Note that if the training video $\mathcal{V}$ contains dynamic entities, we perform a segmentation process to remove these entities before point cloud estimation. Specifically, we first utilize Keye-VL-1.5~\cite{yang2025kwai} to identify dynamic entities and generate corresponding text prompts for each detected entity. Then, we apply ReferDINO~\cite{liang2025referdino} to segment out these dynamic entities, ensuring that the resulting point cloud $\mathcal{S}$ represents only the static components of the scene.

\noindent\textbf{Per-Frame Camera Pose Estimation.} Next, we estimate the camera pose for each frame in $\{\boldsymbol{T}\}^{N} \cup \{\boldsymbol{P}\}^{M} \cup \{\boldsymbol{C}\}^{O}$ using MapAnything~\cite{keetha2025mapanything}. The corresponding per-frame camera poses are denoted as $\{\theta_{\boldsymbol{T}}\}^{N}$, $\{\theta_{\boldsymbol{P}}\}^{M}$ and $\{\theta_{\boldsymbol{C}}\}^{O}$.

\noindent\textbf{View-Specific Scene Point Clouds.} Given the estimated scene point cloud $\mathcal{S}$ and the per-frame camera poses $\{\theta_{\boldsymbol{T}}\}^{N}$, $\{\theta_{\boldsymbol{P}}\}^{M}$ and $\{\theta_{\boldsymbol{C}}\}^{O}$, we apply each camera pose to $\mathcal{S}$ to render the scene from the corresponding viewpoint, as illustrated in Figure~\ref{fig:overview}(a). The resulting view-specific scene point clouds are denoted as $\{\mathcal{S}_{\boldsymbol{T}}\}^{N}$, $\{\mathcal{S}_{\boldsymbol{P}}\}^{M}$ and $\{\mathcal{S}_{\boldsymbol{C}}\}^{O}$, respectively.

\vspace{-1mm}
\subsubsection{Reference Frame Retrieval}
\label{sec:RFS}
The objective of this stage is to select up to $K$ of the most spatially relevant frames from the candidate-frame set $\{\boldsymbol{C}\}^{O}$ as reference frames for the target clip $\{\boldsymbol{T}\}^{N}$. Reference frames are defined as those that exhibit spatial overlap with $\{\boldsymbol{T}\}^{N}$, depicting similar regions or viewpoints within the scene. These frames provide additional spatial cues that enhance geometric consistency during video generation.

To identify these reference frames, we compute spatial correspondence between $\{\boldsymbol{T}\}^{N}$ and $\{\boldsymbol{C}\}^{O}$ using their associated scene point clouds (i.e., $\{\mathcal{S}_{\boldsymbol{T}}\}^{N}$ and $\{\mathcal{S}_{\boldsymbol{C}}\}^{O}$). The detailed retrieval process is illustrated in Figure~\ref{fig:overview}(b) and presented in Algorithm~\ref{alg:retrieval} in the appendix. The retrieved reference-frame set is denoted as $\{\boldsymbol{R}\}^{K} \subset \{\boldsymbol{C}\}^{O}$, where $K$ represents the maximum number of reference frames.

\subsubsection{Architecture}
\label{sec:archi}
Figure~\ref{fig:overview}(c) illustrates the architecture of Spatia, which adopts a multi-modal conditional generation framework. The objective is to generate the target video clip $\{\boldsymbol{T}\}^{N}$ conditioned on the preceding video clip $\{\boldsymbol{P}\}^{M}$, their corresponding scene point clouds $\{\mathcal{S}_{\boldsymbol{T}}\}^{N}$ and $\{\mathcal{S}_{\boldsymbol{P}}\}^{M}$, the retrieved reference frames $\{\boldsymbol{R}\}^{K}$ and the text instruction $\mathcal{T}$.

\noindent\textbf{Token Extraction.} For the video-modality inputs, $\{\boldsymbol{T}\}^{N}$ and $\{\boldsymbol{P}\}^{M}$, we employ the Wan2.2~\cite{wan2025wan} video encoder to convert them into spatio-temporal tokens, denoted as $\boldsymbol{X}_{\boldsymbol{T}}$ and $\boldsymbol{X}_{\boldsymbol{P}}$, respectively. Since this video encoder can also process single-frame images, we use it to encode the image-modality inputs, i.e., the reference frames $\{\boldsymbol{R}\}^{K}$. The resulting token sequence, denoted as $\boldsymbol{X}_{\boldsymbol{R}}$, is obtained by independently encoding each reference frame and concatenating the resulting tokens along the sequence dimension.

As described in Section~\ref{sec:PCE}, $\{\mathcal{S}_{\boldsymbol{T}}\}^{N}$ and $\{\mathcal{S}_{\boldsymbol{P}}\}^{M}$ represent the 3D scene point cloud sequences associated with the target video clip $\{\boldsymbol{T}\}^{N}$ and the preceding video clip $\{\boldsymbol{P}\}^{M}$, respectively. To encode $\{\mathcal{S}_{\boldsymbol{T}}\}^{N}$ and $\{\mathcal{S}_{\boldsymbol{P}}\}^{M}$, each sequence is projected onto the 2D image plane, resulting in a pair of scene projection videos. Missing pixel values in these projection videos are filled with zeros. Both are then processed by the same video encoder, yielding latent representations denoted as $\boldsymbol{X}_{\mathcal{S}_{\boldsymbol{T}}}$ and $\boldsymbol{X}_{\mathcal{S}_{\boldsymbol{P}}}$, respectively. 

At last, to encode the text instruction $\mathcal{T}$, we follow Wan2.2~\cite{wan2025wan} and employ its text encoder to obtain the corresponding text tokens, denoted as $\boldsymbol{X}_{\mathcal{T}}$.

\begin{figure}[!t]
    \centering
\includegraphics[width=0.99\linewidth]{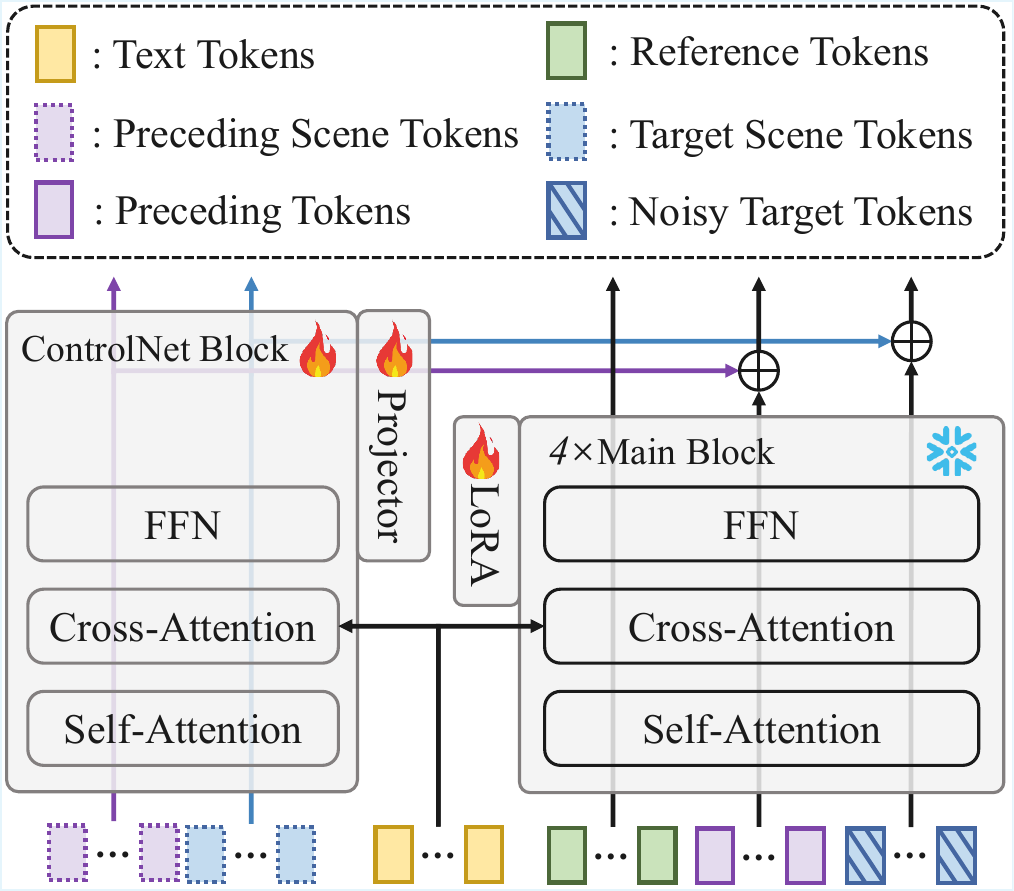}
    \vspace{-2mm}
    \caption{Illustration of a single network block composed of one ControlNet~\cite{zhang2023adding} block operating in parallel with four main blocks. Detailed definitions of all token types are provided in Figure~\ref{fig:overview}.}
    \vspace{-3mm}
    \label{fig:network}
\end{figure}

\noindent\textbf{Network Structure.} Figure~\ref{fig:network} presents the detailed architecture of our network, serving as a complementary illustration to Figure~\ref{fig:overview}(c). We adopt Flow Matching~\cite{lipman2022flow} for model training under multiple conditioning signals—including text tokens $\boldsymbol{X}_{\mathcal{T}}$, reference tokens $\boldsymbol{X}_{\boldsymbol{R}}$, preceding video tokens $\boldsymbol{X}_{\boldsymbol{P}}$, preceding scene video tokens $\boldsymbol{X}_{\mathcal{S}_{\boldsymbol{P}}}$, and target scene video tokens $\boldsymbol{X}_{\mathcal{S}_{\boldsymbol{T}}}$—to guide the generation process from pure noise toward the target video tokens $\boldsymbol{X}_{\boldsymbol{T}}$.

Concretely, given target video tokens $\boldsymbol{X}_{\boldsymbol{T}}$, we first sample $t \in [0, 1]$ from a logit-normal distribution and initialize the noise $\mathbf{x}_0 \sim \mathcal{N}(\mathbf{0}, \mathbf{I})$ following a Gaussian distribution. The intermediate sample $\mathbf{x}_t = (1-t) \mathbf{x}_0 + t \boldsymbol{X}_{\boldsymbol{T}}$ is then obtained via linear interpolation. The model is trained to predict the velocity $\mathbf{u}_t = d\mathbf{x}_t/dt$ by minimizing the mean squared error between the predicted velocity $\mathbf{v}_t$ and the ground-truth velocity $\mathbf{u}_t$:
\begin{equation}
\label{eq:loss}
    \mathcal{L} = \mathbb{E}_{t, \mathbf{x}_0, \boldsymbol{X}_{\boldsymbol{T}}} \left\| \mathbf{v}_t - \mathbf{u}_t \right\|^2.
\end{equation}

Spatia includes 8 network blocks, each containing one ControlNet~\cite{zhang2023adding} block operating in parallel with four main blocks, as illustrated in Figure~\ref{fig:network}. Each main block follows the design of Wan2.2~\cite{wan2025wan}, consisting of a self-attention layer, a cross-attention layer, and an FFN. Each ControlNet block adopts the same architecture but appends a projector—implemented as a simple MLP layer—after the FFN.

\begin{figure}[!t]
    \centering
\includegraphics[width=0.99\linewidth]{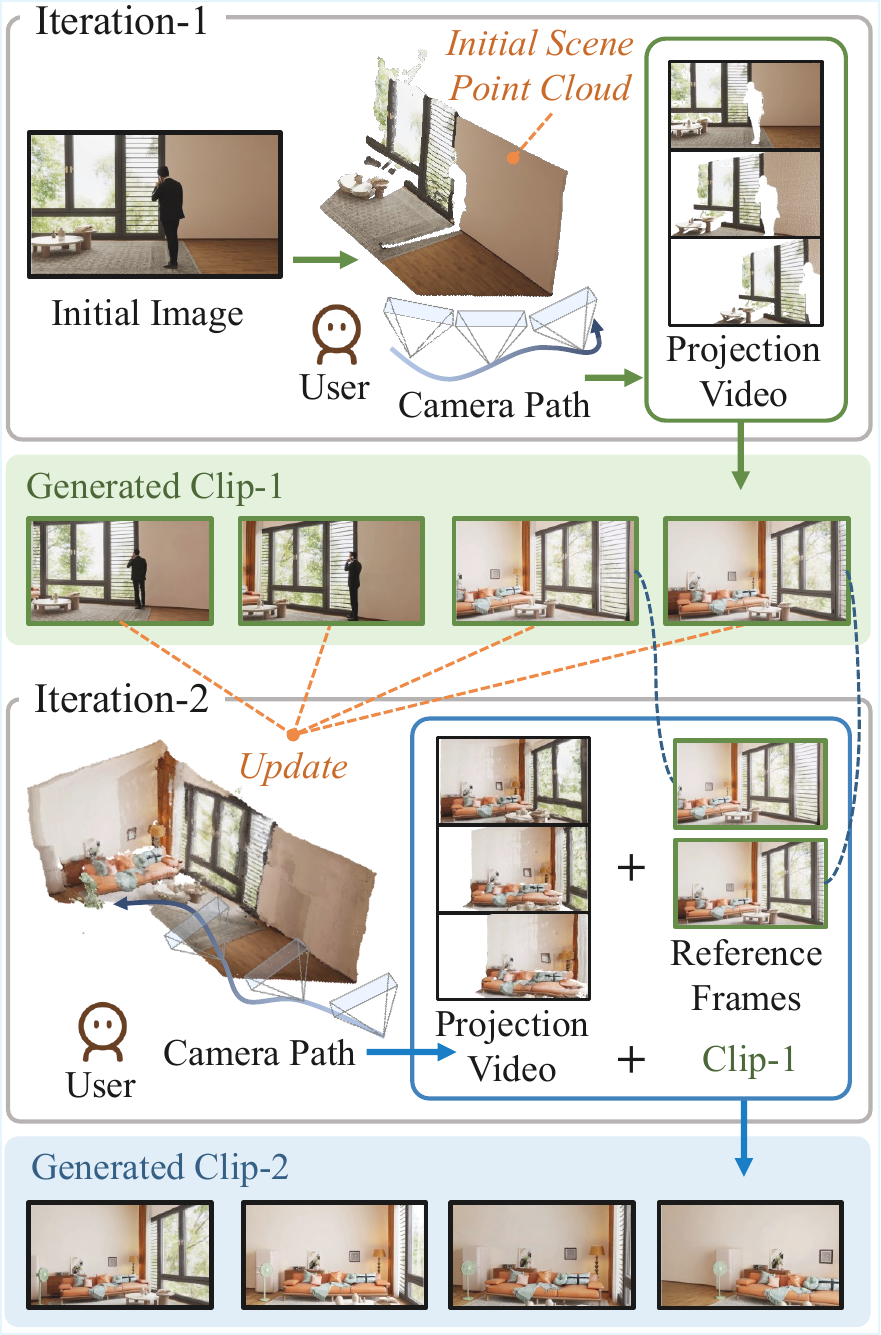}
    \vspace{-3mm}
    \caption{Illustration of the Spatia inference process. At the first iteration, the user provides an initial image, from which Spatia estimates the initial 3D scene point cloud. The user then specifies a text instruction and a camera path based on the estimated scene, producing a projection video along the desired trajectory that conditions the generation of clip-1.
    In subsequent iterations, two steps are performed: (1) Spatia updates the spatial memory (3D scene point cloud) using all previously generated frames via MapAnything~\cite{keetha2025mapanything}; and (2) the user specifies a new text instruction and camera path based on the updated scene. Spatia then takes the reference frames (generated as described in Section~\ref{sec:RFS}), the previously generated clip, and the new projection video as input to produce the next video clip. \textit{Text instructions are omitted.}}
    \vspace{-4mm}
    \label{fig:inference}
\end{figure}

The first ControlNet block processes the concatenation of $\boldsymbol{X}_{\mathcal{S}_{\boldsymbol{P}}}$ and $\boldsymbol{X}_{\mathcal{S}_{\boldsymbol{T}}}$. The resulting outputs are passed to the subsequent ControlNet block and, after projection through a simple MLP layer, the projected features, denoted as $\boldsymbol{X}'_{\mathcal{S}_{\boldsymbol{P}}}$ and $\boldsymbol{X}'_{\mathcal{S}_{\boldsymbol{T}}}$, are also fed into the corresponding main block. Meanwhile, the text tokens $\boldsymbol{X}_{\mathcal{T}}$ are incorporated via the cross-attention layer.

\begin{table*}[t]
    \centering
    \caption{Visual quality comparison on the \textit{WorldScore} benchmark. The final \textbf{Static} and \textbf{Dynamic} world scores are computed by aggregating all relevant metrics. The \textbf{Average} score represents the mean of the static and dynamic world scores. \textit{Static scene generation models} cannot handle dynamic entities, while \textit{foundation video generation models} typically lack persistent memory mechanisms.}
    \vspace{-3mm}
    \setlength{\tabcolsep}{2pt}
    \resizebox{\linewidth}{!}{
        \begin{tabular}{l|ccc|ccc|cccc|ccc}
        \toprule
        Method & \makecell{\textbf{Average} \\ \textbf{Score}} & \makecell{\textbf{Static}\\\textbf{Score}} & \makecell{\textbf{Dynamic}\\\textbf{Score}} & \makecell{Camera \\ Ctrl} & \makecell{Object \\ Ctrl} & \makecell{Content \\ Align} & ~\makecell{3D \\ Const}~ & ~\makecell{Photo \\ Const}~ & ~\makecell{Style \\ Const}~ & \makecell{Subject \\ Quality}~ & \makecell{Motion \\ Acc} & \makecell{Motion \\ Mag} & \makecell{Motion \\ Smooth} \\
        
        \midrule
        \multicolumn{14}{l}{\textit{Static scene generation models}} \\
        WonderJourney~\cite{yu2024wonderjourney} & 54.19 & 63.75 & 44.63 & 84.60 & 37.10 & 35.54 & 80.60 & 79.03 & 62.82 & \underline{66.56} &- &- &-\\
        InvisibleStitch~\cite{engstler2024invisible} & 51.95 & 61.12 & 42.78 & \textbf{93.20} & 36.51 & 29.53 & \textbf{88.51} & \textbf{89.19} & 32.37 & 58.50 & -& - & - \\
         WonderWorld~\cite{yu2024wonderworld} & 61.79 & \underline{72.69} & 50.88 & \underline{92.98} & 51.76 & \underline{71.25} & \underline{86.87} & 85.56 & 70.57 & 49.81 & - & -& - \\
        Voyager~\cite{huang2025voyager} & \underline{66.08} & \textbf{77.62} & 54.53 & 85.95 & \underline{66.92} & 68.92& 81.56 & 85.99 & \textbf{84.89} & \textbf{71.09} &-&- &-\\
        \midrule
        \multicolumn{14}{l}{\textit{Foundation video generation models}} \\
        VideoCrafter2~\cite{chen2024videocrafter2} & 50.03 & 52.57 & 47.49 & 28.92 & 39.07 & \textbf{72.46} & 65.14& 61.85 & 43.79 & 56.74 & 47.12 & 30.40 & 29.39 \\
        EasyAnimate~\cite{xu2024easyanimate} & 52.25 & 52.85 & 51.65 & 26.72 & 54.50 & 50.76 & 67.29 & 47.35 & 73.05 & 50.31 & \underline{75.00} & 31.16 & 40.32 \\
        Allegro~\cite{allegro2024} & 53.64 & 55.31 & 51.97 & 24.84& \underline{57.47} & 51.48 & 70.50 & 69.89 & 65.60 & 47.41 & 54.39 & \textbf{40.28} & 37.81 \\
        CogVideoX-I2V~\cite{yang2024cogvideox} & 60.64 & 62.15 & \underline{59.12} & 38.27& 40.07& 36.73 & 86.21 & 88.12 & \underline{83.22} & 62.44 & 69.56 & 26.42 & 60.15 \\
        Vchitect-2.0~\cite{fan2025vchitect} & 40.38 & 42.28 & 38.47 & 26.55 & 49.54 & 65.75 & 41.53 & 42.30 & 25.69 & 44.58 & 33.59 & \underline{33.81} & 21.31 \\
        LTX-Video~\cite{hacohen2024ltx} & 55.99 & 55.44 & 56.54 & 25.06 & 53.41 & 39.73 & 78.41 & 88.92 & 53.50 & 49.08 & \textbf{76.22} & 29.95 & \underline{71.09} \\
        Wan2.1~\cite{wan2025wan} & 55.21 & 57.56 & 52.85 & 23.53 & 40.32 & 45.44 & 78.74 & 78.36 & 77.18 & 59.38 & 54.27 & 33.26 & 38.05 \\
        \midrule
        Spatia (Ours) & \textbf{69.73} & 72.63 & \textbf{66.82} & 75.66 & 52.32 & 69.95 & 86.40 & \underline{89.10} & 80.09 & 54.86 & 54.83 & 24.75 & \textbf{80.26} \\
        \bottomrule
        \end{tabular}
    }
    \vspace{-4mm}
    \label{tab:worldscore}
\end{table*}

The first main block takes the concatenation of $\boldsymbol{X}_{\boldsymbol{R}}$, $\boldsymbol{X}_{\boldsymbol{P}}$, and $\mathbf{x}_t$ as input. The concatenated tokens are sequentially processed through a stack of layers, including self-attention, cross-attention, and FFN. In the cross-attention layer, the text tokens serve as keys and values, allowing semantic conditioning on textual instructions. This process yields updated features denoted as $\boldsymbol{X}'_{\boldsymbol{R}}$, $\boldsymbol{X}'_{\boldsymbol{P}}$, and $\mathbf{x}'_t$. To integrate scene-level spatial context, the outputs from the associated ControlNet block, $\boldsymbol{X}'_{\mathcal{S}_{\boldsymbol{P}}}$ and $\boldsymbol{X}'_{\mathcal{S}_{\boldsymbol{T}}}$, are fused into the corresponding features via simple addition, resulting in $\boldsymbol{X}'_{\boldsymbol{P}} + \boldsymbol{X}'_{\mathcal{S}_{\boldsymbol{P}}}$ and $\mathbf{x}'_t + \boldsymbol{X}'_{\mathcal{S}_{\boldsymbol{T}}}$. Together with $\boldsymbol{X}'_{\boldsymbol{R}}$, these form the outputs of the block.

\vspace{-1mm}
\subsection{Inference}
\label{sec:inference}
\vspace{-1mm}
Spatia enables iterative user interaction. At each iteration, the user specifies a text instruction and a camera trajectory based on the current 3D scene point cloud to generate a new video clip. The newly generated content, together with previously produced clips, is then used to update the spatial memory (scene point cloud). This iterative process continues, as illustrated in Figure~\ref{fig:inference}.
\vspace{-1mm}
\section{Experiment}
\label{sec:exp}
\vspace{-1mm}

\noindent\textbf{Implementation Details.} Our network backbone is initialized from Wan2.2~\cite{wan2025wan}, containing 5B parameters. Each ControlNet block is initialized from its corresponding main block. The training set consists of two sources: RealEstate~\cite{zhou2018stereo} (40K training videos) and SpatialVID~\cite{wang2025spatialvid} (HD subset, 10K videos), both at 720P resolution. We first train the ControlNet blocks for 8,000 iterations while freezing the main network. Next, we freeze the ControlNet blocks and fine-tune the main blocks using LoRA (rank = 64) for 5,000 iterations. Both stages adopt the AdamW optimizer, with learning rates of 1e-5 and 1e-4, respectively, and a batch size of 64 on $64\times$ AMD MI250 GPUs. By default, the model generates 81 frames for the first (image-conditioned) iteration and 72 frames for each subsequent (clip-conditioned) iteration, conditioned on 9 previously generated frames.

\noindent\textbf{Evaluation.} We evaluate our model from two aspects: (1) visual quality and (2) memory mechanism effectiveness. For (1), we adopt two benchmarks—WorldScore~\cite{duan2025worldscore} and the RealEstate~\cite{zhou2018stereo} test set. The WorldScore benchmark provides 3,000 test samples for text/image-to-video generation and includes a comprehensive suite of metrics to assess both static and dynamic visual quality. For the RealEstate test set, we randomly sample 100 videos, use the first frame of each as the conditioning image, generate corresponding videos, and report PSNR, SSIM, and LPIPS scores against the original videos. For (2), we randomly select 100 samples from the WorldScore benchmark and use each sample’s initial image to generate a closed-loop video—where the camera trajectory brings the final frame back to the initial viewpoint. We then compare the final frame with the initial image using PSNR, SSIM, and LPIPS.

\vspace{-1mm}
\subsection{Main Results}
\label{sec:mainresults}
\vspace{-1mm}
\noindent\textbf{Visual Quality.} 
Using the WorldScore~\cite{duan2025worldscore} benchmark, we evaluate visual quality across three categories of models: (1) \textit{static scene generation models}, which inherently preserve spatial consistency by producing explorable static worlds yet cannot capture motion dynamics; (2) \textit{foundation video generation models}, which generally lack explicit memory mechanisms but effectively generate dynamic content; and (3) \textit{our approach}—a video generation model endowed with spatial memory that integrates dynamic motion generation with long-term spatial coherence. Table~\ref{tab:worldscore} presents the comparison results of \textit{Spatia} against the other two categories of models.

\begin{table}[!t]
    \centering
    \small
    \caption{Evaluation on \textit{RealEstate}. We reproduce the results of all baseline methods using their default configurations and evaluate them on the same test samples to ensure a fair comparison.}
    \vspace{-3mm}
    \begin{tabular}{lccc}
    \toprule
    Method &  PSNR $\uparrow$ &  SSIM $\uparrow$ &  LPIPS $\downarrow$\\
    \midrule
    SEVA \cite{zhou2025stable} & 13.07 & 0.515 & 0.445 \\
    VMem \cite{li2025vmem} & 14.62  &  0.522 & 0.426 \\
    \midrule
    ViewCrafter \cite{yu2024viewcrafter} & 15.78 & 0.580 & 0.396 \\
    FlexWorld \cite{chen2025flexworld} & 16.25  & 0.593   &  0.370 \\
    Voyager \cite{huang2025voyager} & 17.79  & 0.636  & 0.297  \\
    \midrule
    Spatia (Ours) & \textbf{18.58} & \textbf{0.646} & \textbf{0.254}  \\ 
    \bottomrule
    \end{tabular}
    \vspace{-5mm}
    \label{tab:video-realestate}
\end{table}

Meanwhile, for methods that cannot accept control signals from the WorldScore benchmark or have not reported results on it, we evaluate them on the constructed RealEstate~\cite{zhou2018stereo} test set. The results are presented in Table~\ref{tab:video-realestate}. Since RealEstate provides ground-truth videos, we report PSNR, SSIM, and LPIPS by comparing the generated videos against the corresponding ground-truths.

\noindent\textbf{Memory Mechanism Evaluation.} Since few existing methods address video generation with spatial memory, we evaluate our model against scene generation approaches that explicitly maintain spatial memory. The evaluation is conducted on a subset of the WorldScore benchmark containing 100 randomly selected samples. Specifically, we design a \textit{closed-loop} setting where each sample’s initial image is used to generate a video in which the camera trajectory brings the final frame back to the initial viewpoint. We then report PSNR$_C$, SSIM$_C$, and LPIPS$_C$, denoting PSNR, SSIM, and LPIPS between the final frame and the initial image. In addition, we introduce an evaluation metric called Match Accuracy, which measures dense correspondences between the final frame and the initial image—where higher values indicate better spatial alignment. Details are provided in the appendix, and the results are shown in Table~\ref{tab:memory}.

\begin{table}[!t]
    \centering
    \small
    \setlength{\tabcolsep}{2pt}
    \caption{\textit{Memory Mechanism Evaluation on the WorldScore Subset.} Each test sample includes a ground-truth initial image. Using this image, we require the model to generate a \textit{closed-loop} video, where the camera in the final frame returns to the initial viewpoint. We then compute PSNR, SSIM, LPIPS, and Match Accuracy between the final frame and the initial image to evaluate spatial memory consistency.}
    \vspace{-3mm}
    \begin{tabular}{l|cccc}
    \toprule
        Method & PSNR$_{C}$ $\uparrow$  & SSIM$_{C}$ $\uparrow$  & LPIPS$_{C}$ $\downarrow$  & Match Acc $\uparrow$ \\ 
    \midrule
        ViewCrafter~\cite{yu2024viewcrafter} & 14.79 & 0.481 & 0.365 & 0.447  \\ 
        FlexWorld~\cite{chen2025flexworld} & 12.20 & 0.428 & 0.598 & 0.377  \\ 
        Voyager~\cite{huang2025voyager} & 17.66 & 0.540 & 0.380 & 0.507  \\ 
    \midrule
        Spatia (Ours) & \textbf{19.38} & \textbf{0.579} & \textbf{0.213} & \textbf{0.698} \\ 
    \bottomrule
    \end{tabular}
    \vspace{-2mm}
    \label{tab:memory}
\end{table}

\begin{table}[!t]
    \centering
    \small
    \setlength{\tabcolsep}{2pt}
    \caption{Impact of incorporating scene projection videos and reference frames on spatial memory modeling. The ``Camera Control'' metric is adopted from the WorldScore benchmark.}
    \vspace{-3mm}
    \begin{tabular}{cc|cccc}
    \toprule
        \makecell{Scene\\Video} & \makecell{Reference \\ Frames} & \makecell{Camera \\ Control} & PSNR$_{C}$  & SSIM$_{C}$   & LPIPS$_{C}$ \\
    \midrule
        ~ & ~ & 58.81 & 15.55 & 0.444 & 0.379 \\ 
        \checkmark & ~ & 80.13 & 17.18  & 0.500  & 0.295   \\ 
        ~ & \checkmark & 61.38 & 15.64 & 0.444 & 0.393  \\
        \checkmark & \checkmark & \textbf{84.47} & \textbf{19.38} & \textbf{0.579} & \textbf{0.213}  \\
    \bottomrule
    \end{tabular}
    \vspace{-2.5mm}
    \label{tab:module_ablation}    
\end{table}

\vspace{-0.5mm}
\subsection{Ablation Studies}
\vspace{-0.5mm}
Ablation studies related to spatial memory are conducted on the WorldScore subset, with closed-loop videos generated to evaluate spatial memory, as described in Section~\ref{sec:mainresults}. Other studies focusing on visual quality are performed on the RealEstate test set.

\noindent\textbf{Spatial Memory.} Given a camera trajectory, the previously generated frames, and the current spatial memory (i.e., the scene point cloud), we: (1) render a scene projection video along the specified camera path, and (2) retrieve up to $K$ reference frames from prior generations that are spatially correlated with the current trajectory. The rendered scene video and the retrieved reference frames are then provided to our network to facilitate the next-step, memory-aware video generation. The impact of incorporating scene projection videos and reference frames is analyzed in Table~\ref{tab:module_ablation}.

\begin{table}[!t]
    \centering
    \small
    \caption{Effects of using different numbers of reference frames.}
    \vspace{-2mm}
    \begin{tabular}{c|cccc}
    \toprule
        \makecell{\#Reference \\ Frames} & PSNR$_{C}$   & SSIM$_{C}$   & LPIPS$_{C}$   &Match Acc  \\
    \midrule
        1 & 17.50 & 0.537 & 0.284 & 0.592  \\ 
        3 & 17.85 & 0.540 & 0.275  & 0.606  \\ 
        5 & 18.48 & 0.556 & 0.248 & 0.640  \\ 
        7 & \textbf{19.38} & \textbf{0.579} & \textbf{0.213} & \textbf{0.698}   \\
    \bottomrule
    \end{tabular}
    \vspace{-3.5mm}
    \label{tab:ref}    
\end{table}

\noindent\textbf{Reference Frame Number.} Table~\ref{tab:ref} analyzes the effect of varying the number of reference frames $K$. Increasing $K$ provides more spatial memory cues; however, we observe no significant performance improvement when $K > 7$.

\begin{table}[!t]
    \centering
    \small
    \setlength{\tabcolsep}{3pt}
    \caption{Memory mechanisms ensure spatial consistency and preserve visual quality in long-horizon generation.}
    \vspace{-3mm}
        \begin{tabular}{l|c|cccc}
        \toprule
            Method & \makecell{\#Clips} & \makecell{Camera \\ Control} & PSNR$_{C}$  & SSIM$_{C}$   & LPIPS$_{C}$  \\
        \midrule
            \multirowcell{3}[0pt][l]{Wan2.2~\cite{wan2025wan}}
             & 2  & 56.87 & 13.00 & 0.377 & 0.521 \\ 
             & 4  & 46.43 & 11.32 & 0.328 & 0.611  \\ 
             & 6  & 49.97 & 10.74 & 0.310 & 0.644   \\ 
        \midrule
            \multirowcell{3}[0pt][l]{Spatia (Ours)} & 2  & 84.47 & 19.38 &  0.579 & 0.213 \\ 
             & 4  & 83.97 & 18.23 & 0.546 & 0.253  \\ 
             & 6  & 83.41 & 18.04 & 0.541 & 0.259 \\
        \bottomrule
        \end{tabular}
\vspace{-2mm}
    \label{tab:ar}    
\end{table}

\noindent\textbf{Long-Horizon Generation.} Memory is essential for long-horizon generation, ensuring spatial consistency when the camera revisits the same location without visual degradation. To evaluate this, we generate videos of increasing length—2, 4, and 6 clips—under an auto-regressive setting. Each pair of clips moves the camera from left to right and then back to the original viewpoint. We compare our method with Wan2.2 (5B)~\cite{wan2025wan}, where the last frame of each generated clip is used as the starting frame for the next. Table~\ref{tab:ar} summarizes the results, and corresponding visualizations are provided in the appendix.

\begin{table}[!t]
    \centering
    \small
    \caption{Impact of point cloud density on visual quality. Metrics are computed between the generated videos and the ground-truth videos on the RealEstate test set.}
    \vspace{-3mm}
    \begin{tabular}{c|cccc}
    \toprule
        Cube Side Length~(m) & PSNR & SSIM  & LPIPS  \\
    \midrule
        0.01  & \textbf{18.58} & \textbf{0.646} & \textbf{0.254}  \\ 
        0.03  & 17.10 & 0.614 & 0.313  \\ 
        0.05  & 16.35 & 0.596 & 0.349  \\ 
        0.07  & 15.97 & 0.585 & 0.370  \\ 
    \bottomrule
    \end{tabular}
    \vspace{-3mm}
    \label{tab:vocel}    
\end{table}

\noindent\textbf{Scene Point Cloud Density.} We maintain a global scene point cloud as spatial memory. Since the raw point cloud is typically dense, we investigate how its density affects generation quality. Let $d$ denote the side length of each cube used for voxelization. For every cube, we aggregate all points within it to obtain a downsampled point cloud. As shown in Table~\ref{tab:vocel}, increasing $d$ substantially reduces memory storage but leads to visual quality degradation due to the loss of fine-grained spatial guidance.
\vspace{-1mm}
\section{Conclusion}
\vspace{-1mm}
We introduce \textit{Spatia}, a spatial memory–aware video generation framework that enables consistent, long-horizon synthesis. By maintaining an explicit 3D scene point cloud as persistent memory and iteratively updating it during generation, Spatia captures long-term geometric structure that conventional video models cannot preserve. This memory mechanism ensures spatial consistency across revisited locations, supports coherent dynamic content, and enables explicit camera control through 3D-aware conditioning. Extensive experiments demonstrate that Spatia significantly enhances long-horizon consistency while maintaining high visual quality in the generated videos.
\clearpage
\maketitlesupplementary

\section{More Implementation Details}

\noindent\textbf{Reference Frame Retrieval.}
In Section~\ref{sec:RFS} of the main paper, we describe how to select up to $K$ spatially relevant frames (with $K=7$ by default) from the candidate-frame set. The complete procedure is provided in Algorithm~\ref{alg:retrieval}.

\noindent\textbf{Augmentation of Preceding-Frame Latents.}
Spatia conditions on preceding frames to generate future frames, enabling long-horizon video generation. However, while training uses ground-truth preceding frames as conditions, inference relies on model-generated frames, creating a distribution gap between training and inference. To alleviate this mismatch, we introduce a simple augmentation strategy for preceding-frame latents during training. Specifically, we sample a timestep $t_{\text{aug}} \in [0, 50]$ from a low-noise interval using the same noise scheduler as in Flow Matching training, and add the corresponding noise to the clean preceding-frame latents. The resulting augmented latents are then used as the conditioning inputs in place of the clean latents.

\begin{algorithm}[!t]
\caption{Reference Frame Retrieval}
\label{alg:retrieval}
\begin{algorithmic}[1]
\Require Target frames $\{\boldsymbol{T}\}^{N}$, candidate frames $\{\boldsymbol{C}\}^{O}$, view-specific scene point clouds $\{\mathcal{S}_{\boldsymbol{T}}\}^{N}$ and $\{\mathcal{S}_{\boldsymbol{C}}\}^{O}$, threshold $\epsilon$, maximum number of reference frames $K$
\Ensure Retrieved reference-frame set $\{\boldsymbol{R}\}$

\State Initialize $\{\boldsymbol{R}\} \gets \emptyset$

\For{each target frame $\boldsymbol{T}_i \in \{\boldsymbol{T}\}^{N}$}
    \If{$i \bmod K \neq 0$}
        \State \textbf{break} \Comment{Operate every $K$ frames.}
    \EndIf
    \State Initialize $s \gets 0$ \Comment{Maximal spatial overlap score.}
    \State Initialize $\boldsymbol{\hat{R}} \gets \emptyset$ \Comment{Empty reference frame.}
    \State Identify the scene map $\mathcal{S}_{\boldsymbol{T}_i} \in \{\mathcal{S}_{\boldsymbol{T}}\}^{N}$
    \For{each candidate frame $\boldsymbol{C}_j \in \{\boldsymbol{C}\}^{O}$}
    \State Identify the scene map $\mathcal{S}_{\boldsymbol{C}_j} \in \{\mathcal{S}_{\boldsymbol{C}}\}^{O}$  
    \State $s(\boldsymbol{T}_i, \boldsymbol{C}_j) \gets \Call{SpatialOverlap}{\mathcal{S}_{\boldsymbol{T}_i}, \mathcal{S}_{\boldsymbol{C}_j}}$
    \If{$s(\boldsymbol{T}_i, \boldsymbol{C}_j) > s$}
        \State $s\gets\ s(\boldsymbol{T}_i, \boldsymbol{C}_j)$
        \State $\boldsymbol{\hat{R}} \gets\ \boldsymbol{C}_j$
    \EndIf
    \EndFor
    \If{$s > \epsilon$}
        \State $\{\boldsymbol{R}\} \gets \{\boldsymbol{R}\} \cup \boldsymbol{\hat{R}}$
    \EndIf
\EndFor

\State \Return $\{\boldsymbol{R}\}$

\Function{SpatialOverlap}{$x, y$}
  \State $y' \gets \text{Register}(y,x)$ \Comment{Register $y$ to $x$ space.}
  \State $s \gets \text{3DIoU}(x,y')$
  \State \Return $s$
\EndFunction
\end{algorithmic}
\end{algorithm}

\noindent\textbf{Match Accuracy.}
In Tables~\ref{tab:memory} and \ref{tab:ref} of the main paper, we include \textit{Match Accuracy} as an additional metric to assess the effectiveness of the memory mechanism in closed-loop video generation, where the last frame is expected to reproduce spatially similar content to the initial frame. \textit{Match Accuracy} quantifies the structural and spatial correspondence between two frames. In our implementation, we use RoMa~\cite{edstedt2024roma}, a robust dense feature-matching algorithm, to estimate correspondences between the first frame $\boldsymbol{I}_{\text{first}}$ and the last frame $\boldsymbol{I}_{\text{last}}$. After obtaining the correspondence map, we discard low-confidence matches and count the remaining high-confidence correspondences as the number of valid matches. To ensure comparability across scenes, the final match accuracy is normalized by the number of high-confidence self-matches obtained by matching $\boldsymbol{I}_{\text{first}}$ with itself.

\noindent\textbf{Dynamic-Static Disentanglement in the Inference Stage.}
Our model supports generating videos that contain dynamic entities while maintaining a spatial memory representing the static scene. During inference, to strictly enforce dynamic–static disentanglement, we first apply SAM2~\cite{ravi2024sam2} to track and segment dynamic entities in the initial conditioning image or previously generated video clips, and record their segmentation masks. These masks are then used to exclude dynamic regions when updating the spatial memory (i.e., the scene point cloud) with MapAnything~\cite{keetha2025mapanything}.

\section{Visualization}
\label{sec:visual}

\begin{figure*}[!t]
    \centering
    \includegraphics[width=0.99\linewidth]{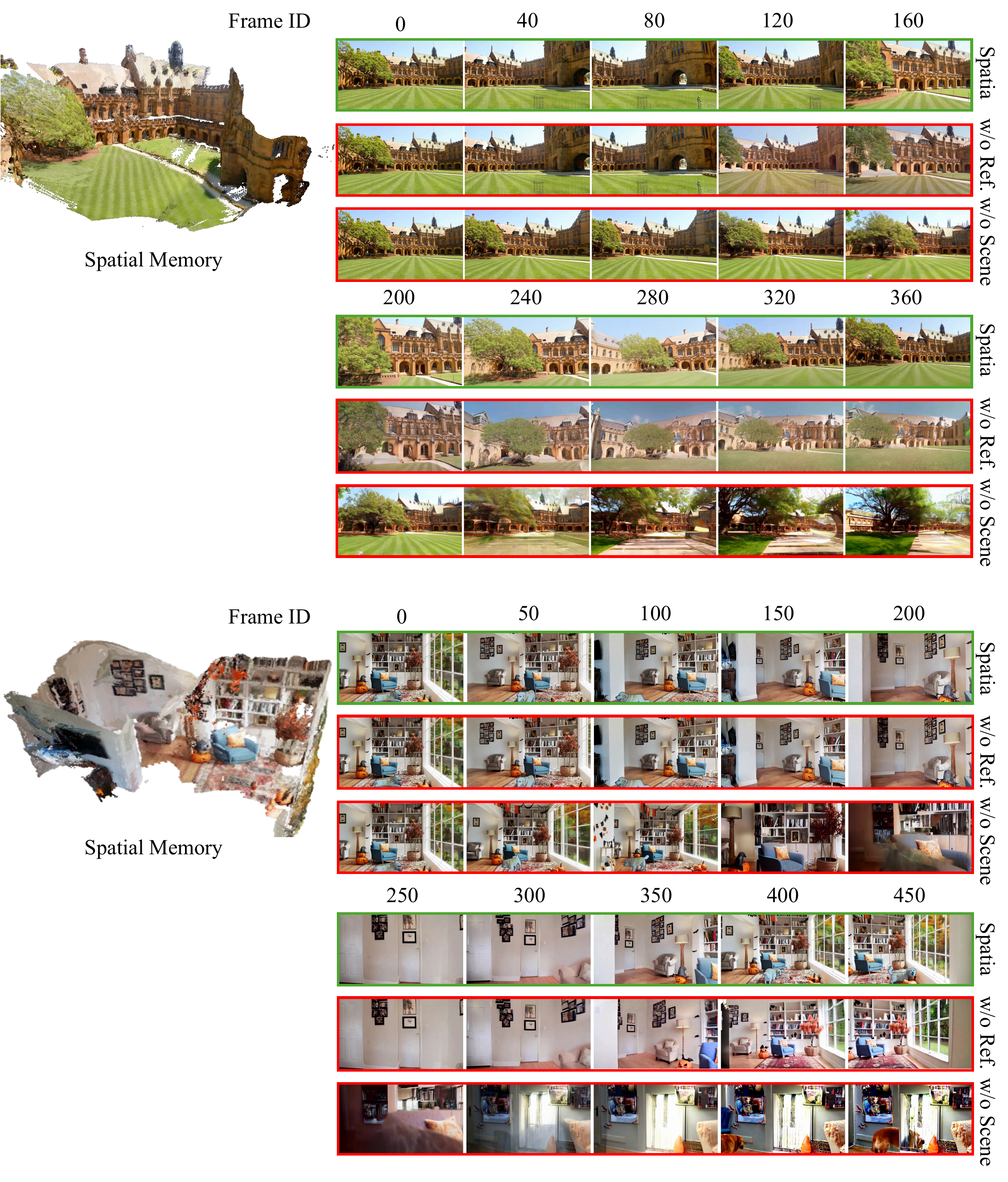}

    \caption{Qualitative comparison of three variants for long-horizon video generation: (1) our default model Spatia, (2) a variant using only scene videos without reference frames, and (3) a variant using reference frames but no scene videos. The spatial memories shown in the figure are generated by Spatia.}

    \label{fig:ablation}
\end{figure*}

\noindent\textbf{Qualitative Study on Spatial Memory in Long-Horizon Generation.}
In Table~\ref{tab:module_ablation} of the main paper, we quantitatively study two key factors for enabling spatial memory and achieving spatially consistent long-horizon generation: (1) the use of reference frames and (2) the use of scene videos. Figure~\ref{fig:ablation} presents a qualitative comparison among three variants: (1) our default model incorporating both components, (2) a model that uses only scene videos without reference frames, and (3) a model that uses reference frames but excludes scene videos. As shown, our full model substantially outperforms both ablated variants, successfully preserving global scene consistency and structural integrity over long temporal sequences, while the baselines exhibit pronounced geometric drift.

\begin{figure*}[!t]
    \centering
    \includegraphics[width=0.95\linewidth]{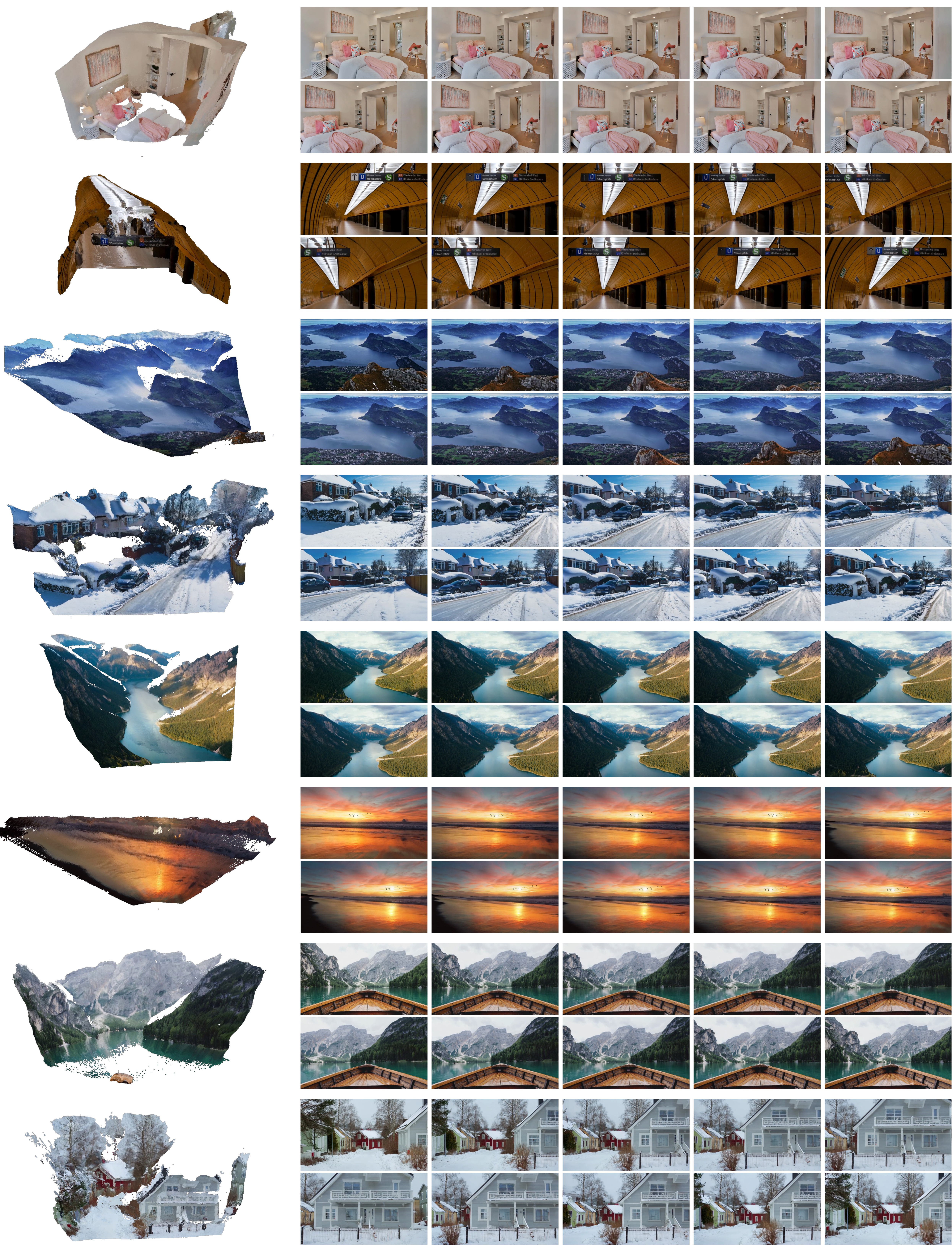}

    \caption{Visualization of closed-loop video generation. The camera follows a trajectory that returns to its initial viewpoint, enabling direct comparison between the first and final frames to evaluate the effectiveness of the spatial memory mechanism.}

    \label{fig:closed_loop}
\end{figure*}

\noindent\textbf{Closed-Loop Generation}
Figure~\ref{fig:closed_loop} shows visualizations of closed-loop video generation. In these examples, the camera follows a trajectory that returns to the initial viewpoint at the end of the sequence. This setup enables direct evaluation of both visual and geometric consistency by examining whether the final frame spatially aligns with the first frame, thereby validating the effectiveness of our spatial memory in preserving global scene structure.

\begin{figure*}[!t]
    \centering
\includegraphics[width=0.93\linewidth]{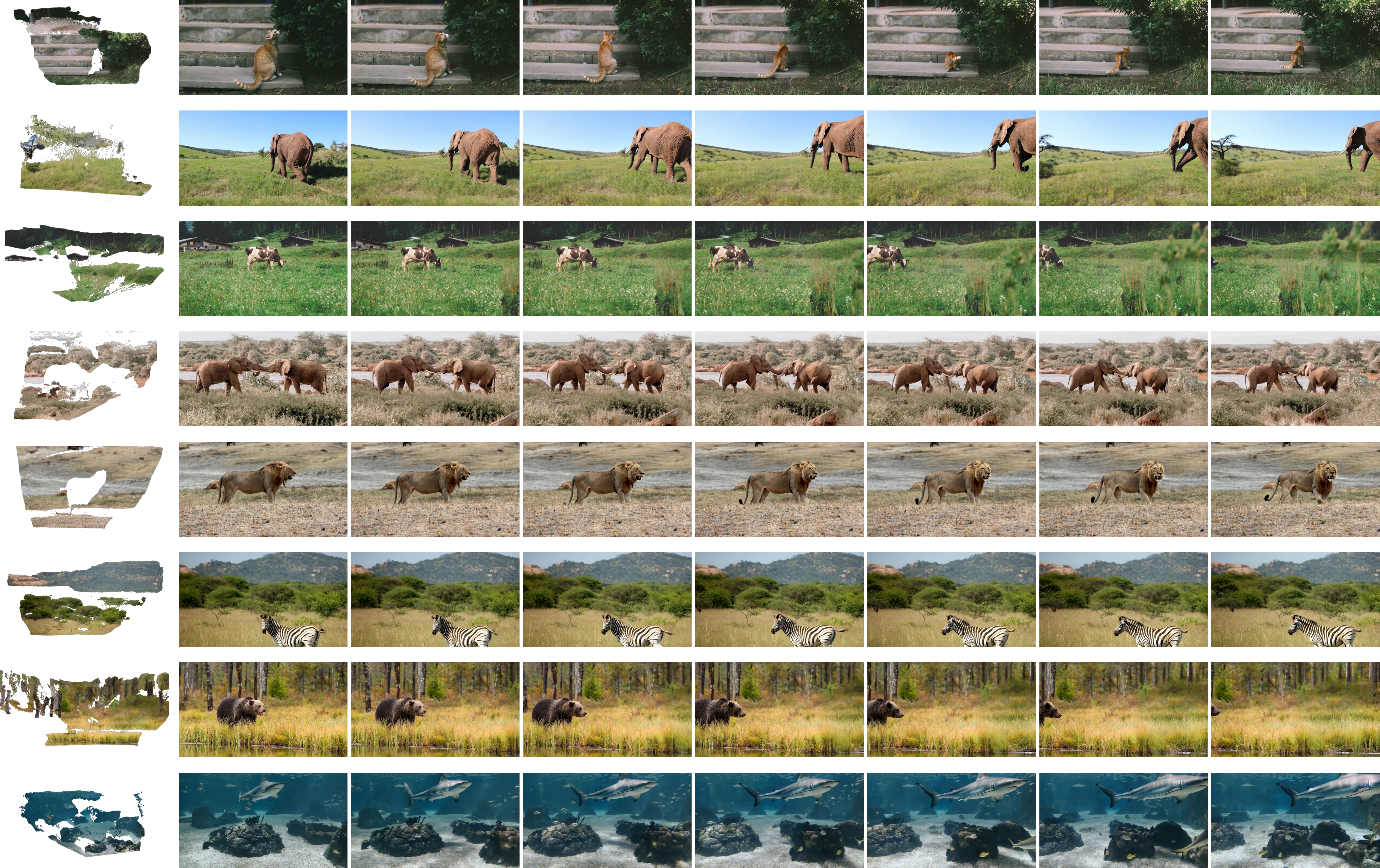}
    \vspace{-1mm}
    \caption{Visualizations of dynamic–static disentanglement. Our model maintains a spatial memory containing only the static scene point cloud while generating videos that include dynamic entities acting within the same scenes.}
\vspace{-2mm}
    \label{fig:dynamic}
\end{figure*}

\noindent\textbf{Generation of Dynamic Entities while Maintaining Static Scenes.}
Our model supports dynamic–static disentanglement by representing only the static scene in the spatial memory. This is accomplished by removing dynamic entities from the estimated scene point cloud, which is used as the spatial memory, while the original videos containing dynamic entities are used as training targets. Figure~\ref{fig:dynamic} illustrates several examples showing the static-only spatial memory alongside the corresponding generated videos, where dynamic entities perform actions within the same scenes.

\begin{figure*}[!h]
    \centering
    \includegraphics[width=0.93\linewidth]{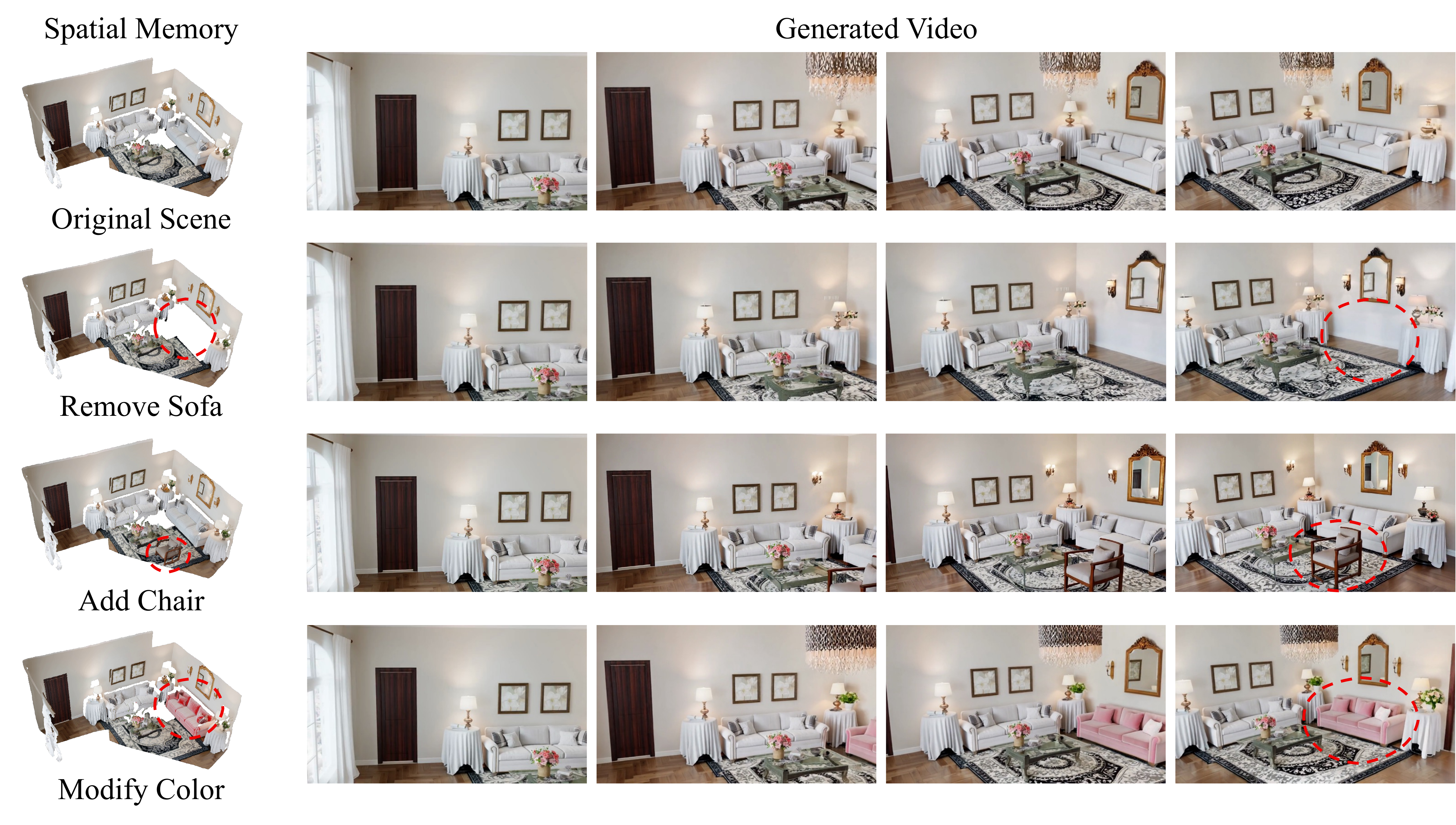}
    \vspace{-3mm}
    \caption{Demonstration of 3D-aware interactive editing. By directly modifying the spatial memory (i.e., the scene point cloud), users can achieve geometrically precise edits in the generated videos, such as removing an object (2nd row), adding a new object (3rd row), or altering object attributes (4th row).}
    \label{fig:edit}
\end{figure*}

\noindent\textbf{3D-Aware Interactive Editing}
Maintaining a scene point cloud as spatial memory and conditioning on it during video generation also enables 3D-aware interactive editing. As shown in Figure~\ref{fig:edit}, manipulating the estimated scene point cloud—such as removing objects, adding new ones, or modifying object colors—leads to corresponding and accurate changes in the generated videos.

\clearpage
{
    \small
    \bibliographystyle{ieeenat_fullname}
    \bibliography{main}

@String(CVPR= {IEEE Conf. Comput. Vis. Pattern Recog.})

@String(ICCV= {Int. Conf. Comput. Vis.})

@String(ECCV= {Eur. Conf. Comput. Vis.})

@String(CVPR  = {CVPR})

@String(ICCV  = {ICCV})

@String(ECCV  = {ECCV})

@techreport{anthropic2024claude3,
  title        = {Claude 3 Model Card},
  author       = {Anthropic},
  institution  = {Anthropic PBC},
  year         = {2024},
  url          = {https://www-cdn.anthropic.com/de8ba9b01c9ab7cbabf5c33b80b7bbc618857627/Model_Card_Claude_3.pdf},
}

@inproceedings{shenvideovla,
  title={VideoVLA: Video Generators Can Be Generalizable Robot Manipulators},
  author={Shen, Yichao and Wei, Fangyun and Du, Zhiying and Liang, Yaobo and Lu, Yan and Yang, Jiaolong and Zheng, Nanning and Guo, Baining},
  booktitle={The Thirty-ninth Annual Conference on Neural Information Processing Systems}
}

@article{achiam2023gpt,
  title={Gpt-4 technical report},
  author={Achiam, Josh and Adler, Steven and Agarwal, Sandhini and Ahmad, Lama and Akkaya, Ilge and Aleman, Florencia Leoni and Almeida, Diogo and Altenschmidt, Janko and Altman, Sam and Anadkat, Shyamal and others},
  journal={arXiv preprint arXiv:2303.08774},
  year={2023}
}

@article{team2024gemini,
  title={Gemini 1.5: Unlocking multimodal understanding across millions of tokens of context},
  author={Team, Gemini and Georgiev, Petko and Lei, Ving Ian and Burnell, Ryan and Bai, Libin and Gulati, Anmol and Tanzer, Garrett and Vincent, Damien and Pan, Zhufeng and Wang, Shibo and others},
  journal={arXiv preprint arXiv:2403.05530},
  year={2024}
}

@misc{google2025gemini2flash,
  author       = {Google DeepMind},
  title        = {Gemini 2.0 Flash: A Multimodal Model with 1 Million Token Context Window},
  howpublished = {\url{https://cloud.google.com/vertex-ai/generative-ai/docs/models/gemini/2-0-flash}},
  year         = {2025},
}

@techreport{anthropic2025claude4,
  title       = {Claude 4 Model Card (Claude Opus 4 \& Sonnet 4)},
  author      = {Anthropic},
  institution = {Anthropic PBC},
  month       = {May},
  year        = {2025},
  howpublished= {\url{https://www-cdn.anthropic.com/4263b940cabb546aa0e3283f35b686f4f3b2ff47.pdf}},
}

@misc{openai2025gpt41,
  author       = {OpenAI},
  title        = {Introducing GPT-4.1 in the API},
  howpublished = {\url{https://openai.com/index/gpt-4-1/}},
  year         = {2025}
}

@misc{meta2025llama4,
  author = {{Meta-AI}},
  title = {The Llama 4 herd: The beginning of a new era of natively multimodal AI innovation},
  howpublished = {Meta AI Blog},
  year = {2025},
  month = {April},
  url = {https://ai.meta.com/blog/llama-4-multimodal-intelligence/},
}

@article{brown2020language,
  title={Language models are few-shot learners},
  author={Brown, Tom and Mann, Benjamin and Ryder, Nick and Subbiah, Melanie and Kaplan, Jared D and Dhariwal, Prafulla and Neelakantan, Arvind and Shyam, Pranav and Sastry, Girish and Askell, Amanda and others},
  journal={Advances in neural information processing systems},
  volume={33},
  pages={1877--1901},
  year={2020}
}

@article{wang2025pi,
  title={Pi3: Scalable Permutation-Equivariant Visual Geometry Learning},
  author={Wang, Yifan and Zhou, Jianjun and Zhu, Haoyi and Chang, Wenzheng and Zhou, Yang and Li, Zizun and Chen, Junyi and Pang, Jiangmiao and Shen, Chunhua and He, Tong},
  journal={arXiv e-prints},
  pages={arXiv--2507},
  year={2025}
}

@article{zhou2025stable,
  title={Stable virtual camera: Generative view synthesis with diffusion models},
  author={Zhou, Jensen and Gao, Hang and Voleti, Vikram and Vasishta, Aaryaman and Yao, Chun-Han and Boss, Mark and Torr, Philip and Rupprecht, Christian and Jampani, Varun},
  journal={arXiv preprint arXiv:2503.14489},
  year={2025}
}

@inproceedings{devlin2019bert,
  title={Bert: Pre-training of deep bidirectional transformers for language understanding},
  author={Devlin, Jacob and Chang, Ming-Wei and Lee, Kenton and Toutanova, Kristina},
  booktitle={Proceedings of the 2019 conference of the North American chapter of the association for computational linguistics: human language technologies, volume 1 (long and short papers)},
  pages={4171--4186},
  year={2019}
}

@article{ding2024longrope,
  title={Longrope: Extending llm context window beyond 2 million tokens},
  author={Ding, Yiran and Zhang, Li Lyna and Zhang, Chengruidong and Xu, Yuanyuan and Shang, Ning and Xu, Jiahang and Yang, Fan and Yang, Mao},
  journal={arXiv preprint arXiv:2402.13753},
  year={2024}
}

@inproceedings{kwon2023efficient,
  title={Efficient memory management for large language model serving with pagedattention},
  author={Kwon, Woosuk and Li, Zhuohan and Zhuang, Siyuan and Sheng, Ying and Zheng, Lianmin and Yu, Cody Hao and Gonzalez, Joseph and Zhang, Hao and Stoica, Ion},
  booktitle={Proceedings of the 29th symposium on operating systems principles},
  pages={611--626},
  year={2023}
}

@article{li2024snapkv,
  title={Snapkv: Llm knows what you are looking for before generation},
  author={Li, Yuhong and Huang, Yingbing and Yang, Bowen and Venkitesh, Bharat and Locatelli, Acyr and Ye, Hanchen and Cai, Tianle and Lewis, Patrick and Chen, Deming},
  journal={Advances in Neural Information Processing Systems},
  volume={37},
  pages={22947--22970},
  year={2024}
}

@article{liu2023ring,
  title={Ring attention with blockwise transformers for near-infinite context},
  author={Liu, Hao and Zaharia, Matei and Abbeel, Pieter},
  journal={arXiv preprint arXiv:2310.01889},
  year={2023}
}

@article{radford2018improving,
  title={Improving language understanding by generative pre-training},
  author={Radford, Alec and Narasimhan, Karthik and Salimans, Tim and Sutskever, Ilya and others},
  year={2018},
  publisher={San Francisco, CA, USA}
}

@article{radford2019language,
  title={Language models are unsupervised multitask learners},
  author={Radford, Alec and Wu, Jeffrey and Child, Rewon and Luan, David and Amodei, Dario and Sutskever, Ilya and others},
  journal={OpenAI blog},
  volume={1},
  number={8},
  pages={9},
  year={2019}
}

@article{raffel2020exploring,
  title={Exploring the limits of transfer learning with a unified text-to-text transformer},
  author={Raffel, Colin and Shazeer, Noam and Roberts, Adam and Lee, Katherine and Narang, Sharan and Matena, Michael and Zhou, Yanqi and Li, Wei and Liu, Peter J},
  journal={Journal of machine learning research},
  volume={21},
  number={140},
  pages={1--67},
  year={2020}
}

@article{xiong2023effective,
  title={Effective long-context scaling of foundation models},
  author={Xiong, Wenhan and Liu, Jingyu and Molybog, Igor and Zhang, Hejia and Bhargava, Prajjwal and Hou, Rui and Martin, Louis and Rungta, Rashi and Sankararaman, Karthik Abinav and Oguz, Barlas and others},
  journal={arXiv preprint arXiv:2309.16039},
  year={2023}
}

@article{zhang2023h2o,
  title={H2o: Heavy-hitter oracle for efficient generative inference of large language models},
  author={Zhang, Zhenyu and Sheng, Ying and Zhou, Tianyi and Chen, Tianlong and Zheng, Lianmin and Cai, Ruisi and Song, Zhao and Tian, Yuandong and R{\'e}, Christopher and Barrett, Clark and others},
  journal={Advances in Neural Information Processing Systems},
  volume={36},
  pages={34661--34710},
  year={2023}
}

@article{yu2024viewcrafter,
  title={Viewcrafter: Taming video diffusion models for high-fidelity novel view synthesis},
  author={Yu, Wangbo and Xing, Jinbo and Yuan, Li and Hu, Wenbo and Li, Xiaoyu and Huang, Zhipeng and Gao, Xiangjun and Wong, Tien-Tsin and Shan, Ying and Tian, Yonghong},
  journal={arXiv preprint arXiv:2409.02048},
  year={2024}
}

@article{liu2025infinitystar,
  title={InfinityStar: Unified Spacetime AutoRegressive Modeling for Visual Generation},
  author={Liu, Jinlai and Han, Jian and Yan, Bin and Wu, Hui and Zhu, Fengda and Wang, Xing and Jiang, Yi and Peng, Bingyue and Yuan, Zehuan},
  journal={arXiv preprint arXiv:2511.04675},
  year={2025}
}

@article{chen2025flexworld,
  title={FlexWorld: Progressively expanding 3D scenes for flexiable-view synthesis},
  author={Chen, Luxi and Zhou, Zihan and Zhao, Min and Wang, Yikai and Zhang, Ge and Huang, Wenhao and Sun, Hao and Wen, Ji-Rong and Li, Chongxuan},
  journal={arXiv preprint arXiv:2503.13265},
  year={2025}
}

@inproceedings{ma2025you,
  title={You see it, you got it: Learning 3d creation on pose-free videos at scale},
  author={Ma, Baorui and Gao, Huachen and Deng, Haoge and Luo, Zhengxiong and Huang, Tiejun and Tang, Lulu and Wang, Xinlong},
  booktitle={Proceedings of the Computer Vision and Pattern Recognition Conference},
  pages={2016--2029},
  year={2025}
}

@article{yu2025context,
  title={Context as memory: Scene-consistent interactive long video generation with memory retrieval},
  author={Yu, Jiwen and Bai, Jianhong and Qin, Yiran and Liu, Quande and Wang, Xintao and Wan, Pengfei and Zhang, Di and Liu, Xihui},
  journal={arXiv preprint arXiv:2506.03141},
  year={2025}
}

@article{huang2025voyager,
  title={Voyager: Long-Range and World-Consistent Video Diffusion for Explorable 3D Scene Generation},
  author={Huang, Tianyu and Zheng, Wangguandong and Wang, Tengfei and Liu, Yuhao and Wang, Zhenwei and Wu, Junta and Jiang, Jie and Li, Hui and Lau, Rynson WH and Zuo, Wangmeng and others},
  journal={arXiv preprint arXiv:2506.04225},
  year={2025}
}

@article{li2025vmem,
  title={VMem: Consistent Interactive Video Scene Generation with Surfel-Indexed View Memory},
  author={Li, Runjia and Torr, Philip and Vedaldi, Andrea and Jakab, Tomas},
  journal={arXiv preprint arXiv:2506.18903},
  year={2025}
}

@article{blattmann2023stable,
  title={Stable video diffusion: Scaling latent video diffusion models to large datasets},
  author={Blattmann, Andreas and Dockhorn, Tim and Kulal, Sumith and Mendelevitch, Daniel and Kilian, Maciej and Lorenz, Dominik and Levi, Yam and English, Zion and Voleti, Vikram and Letts, Adam and others},
  journal={arXiv preprint arXiv:2311.15127},
  year={2023}
}

@article{chen2023videocrafter1,
  title={Videocrafter1: Open diffusion models for high-quality video generation},
  author={Chen, Haoxin and Xia, Menghan and He, Yingqing and Zhang, Yong and Cun, Xiaodong and Yang, Shaoshu and Xing, Jinbo and Liu, Yaofang and Chen, Qifeng and Wang, Xintao and others},
  journal={arXiv preprint arXiv:2310.19512},
  year={2023}
}

@inproceedings{chen2024videocrafter2,
  title={Videocrafter2: Overcoming data limitations for high-quality video diffusion models},
  author={Chen, Haoxin and Zhang, Yong and Cun, Xiaodong and Xia, Menghan and Wang, Xintao and Weng, Chao and Shan, Ying},
  booktitle={Proceedings of the IEEE/CVF Conference on Computer Vision and Pattern Recognition},
  pages={7310--7320},
  year={2024}
}

@inproceedings{zeng2024make,
  title={Make pixels dance: High-dynamic video generation},
  author={Zeng, Yan and Wei, Guoqiang and Zheng, Jiani and Zou, Jiaxin and Wei, Yang and Zhang, Yuchen and Li, Hang},
  booktitle={Proceedings of the IEEE/CVF Conference on Computer Vision and Pattern Recognition},
  pages={8850--8860},
  year={2024}
}

@inproceedings{peebles2023scalable,
  title={Scalable diffusion models with transformers},
  author={Peebles, William and Xie, Saining},
  booktitle={Proceedings of the IEEE/CVF international conference on computer vision},
  pages={4195--4205},
  year={2023}
}

@inproceedings{esser2024scaling,
  title={Scaling rectified flow transformers for high-resolution image synthesis},
  author={Esser, Patrick and Kulal, Sumith and Blattmann, Andreas and Entezari, Rahim and M{\"u}ller, Jonas and Saini, Harry and Levi, Yam and Lorenz, Dominik and Sauer, Axel and Boesel, Frederic and others},
  booktitle={Forty-first international conference on machine learning},
  year={2024}
}

@article{yang2024cogvideox,
  title={Cogvideox: Text-to-video diffusion models with an expert transformer},
  author={Yang, Zhuoyi and Teng, Jiayan and Zheng, Wendi and Ding, Ming and Huang, Shiyu and Xu, Jiazheng and Yang, Yuanming and Hong, Wenyi and Zhang, Xiaohan and Feng, Guanyu and others},
  journal={arXiv preprint arXiv:2408.06072},
  year={2024}
}

@article{kong2024hunyuanvideo,
  title={Hunyuanvideo: A systematic framework for large video generative models},
  author={Kong, Weijie and Tian, Qi and Zhang, Zijian and Min, Rox and Dai, Zuozhuo and Zhou, Jin and Xiong, Jiangfeng and Li, Xin and Wu, Bo and Zhang, Jianwei and others},
  journal={arXiv preprint arXiv:2412.03603},
  year={2024}
}

@article{wan2025wan,
      title={Wan: Open and Advanced Large-Scale Video Generative Models}, 
      author={Team Wan and Ang Wang and Baole Ai and Bin Wen and Chaojie Mao and Chen-Wei Xie and Di Chen and Feiwu Yu and Haiming Zhao and Jianxiao Yang and Jianyuan Zeng and Jiayu Wang and Jingfeng Zhang and Jingren Zhou and Jinkai Wang and Jixuan Chen and Kai Zhu and Kang Zhao and Keyu Yan and Lianghua Huang and Mengyang Feng and Ningyi Zhang and Pandeng Li and Pingyu Wu and Ruihang Chu and Ruili Feng and Shiwei Zhang and Siyang Sun and Tao Fang and Tianxing Wang and Tianyi Gui and Tingyu Weng and Tong Shen and Wei Lin and Wei Wang and Wei Wang and Wenmeng Zhou and Wente Wang and Wenting Shen and Wenyuan Yu and Xianzhong Shi and Xiaoming Huang and Xin Xu and Yan Kou and Yangyu Lv and Yifei Li and Yijing Liu and Yiming Wang and Yingya Zhang and Yitong Huang and Yong Li and You Wu and Yu Liu and Yulin Pan and Yun Zheng and Yuntao Hong and Yupeng Shi and Yutong Feng and Zeyinzi Jiang and Zhen Han and Zhi-Fan Wu and Ziyu Liu},
      journal = {arXiv preprint arXiv:2503.20314},
      year={2025}
}

@article{team2025longcat,
  title={LongCat-Video Technical Report},
  author={Team, Meituan LongCat and Cai, Xunliang and Huang, Qilong and Kang, Zhuoliang and Li, Hongyu and Liang, Shijun and Ma, Liya and Ren, Siyu and Wei, Xiaoming and Xie, Rixu and others},
  journal={arXiv preprint arXiv:2510.22200},
  year={2025}
}

@article{hacohen2024ltx,
  title={Ltx-video: Realtime video latent diffusion},
  author={HaCohen, Yoav and Chiprut, Nisan and Brazowski, Benny and Shalem, Daniel and Moshe, Dudu and Richardson, Eitan and Levin, Eran and Shiran, Guy and Zabari, Nir and Gordon, Ori and others},
  journal={arXiv preprint arXiv:2501.00103},
  year={2024}
}

@misc{veo,
  title = {Veo},
  author = {Google},
  howpublished = {\url{https://deepmind.google/models/veo/}},
  year = {2024},
}

@misc{sora,
  title = {Sora},
  author = {OpenAI},
  howpublished = {\url{https://openai.com/sora/}},
  year = {2024},
}

@misc{Hailuo,
  title = {Hailuo},
  author = {MiniMax},
  howpublished = {\url{https://hailuoai.video/}},
  year = {2024},
}

@misc{Kling,
  title = {Kling},
  author = {Kuaishou},
  howpublished = {\url{https://klingai.com}},
  year = {2024},
}

@article{sun2025virtual,
  title={From Virtual Games to Real-World Play},
  author={Sun, Wenqiang and Wei, Fangyun and Zhao, Jinjing and Chen, Xi and Chen, Zilong and Zhang, Hongyang and Zhang, Jun and Lu, Yan},
  journal={arXiv preprint arXiv:2506.18901},
  year={2025}
}

@article{gao2025seedance,
  title={Seedance 1.0: Exploring the Boundaries of Video Generation Models},
  author={Gao, Yu and Guo, Haoyuan and Hoang, Tuyen and Huang, Weilin and Jiang, Lu and Kong, Fangyuan and Li, Huixia and Li, Jiashi and Li, Liang and Li, Xiaojie and others},
  journal={arXiv preprint arXiv:2506.09113},
  year={2025}
}

@article{zheng2024open1,
  title={Open-sora: Democratizing efficient video production for all},
  author={Zheng, Zangwei and Peng, Xiangyu and Yang, Tianji and Shen, Chenhui and Li, Shenggui and Liu, Hongxin and Zhou, Yukun and Li, Tianyi and You, Yang},
  journal={arXiv preprint arXiv:2412.20404},
  year={2024}
}

@article{lin2024open_plan,
  title={Open-sora plan: Open-source large video generation model},
  author={Lin, Bin and Ge, Yunyang and Cheng, Xinhua and Li, Zongjian and Zhu, Bin and Wang, Shaodong and He, Xianyi and Ye, Yang and Yuan, Shenghai and Chen, Liuhan and others},
  journal={arXiv preprint arXiv:2412.00131},
  year={2024}
}

@article{peng2025open2,
  title={Open-sora 2.0: Training a commercial-level video generation model in \$200 k},
  author={Peng, Xiangyu and Zheng, Zangwei and Shen, Chenhui and Young, Tom and Guo, Xinying and Wang, Binluo and Xu, Hang and Liu, Hongxin and Jiang, Mingyan and Li, Wenjun and others},
  journal={arXiv preprint arXiv:2503.09642},
  year={2025}
}

@article{ma2025step,
  title={Step-video-t2v technical report: The practice, challenges, and future of video foundation model},
  author={Ma, Guoqing and Huang, Haoyang and Yan, Kun and Chen, Liangyu and Duan, Nan and Yin, Shengming and Wan, Changyi and Ming, Ranchen and Song, Xiaoniu and Chen, Xing and others},
  journal={arXiv preprint arXiv:2502.10248},
  year={2025}
}

@article{kim2024fifo,
  title={Fifo-diffusion: Generating infinite videos from text without training},
  author={Kim, Jihwan and Kang, Junoh and Choi, Jinyoung and Han, Bohyung},
  journal={arXiv preprint arXiv:2405.11473},
  year={2024}
}

@article{jin2024pyramidal,
  title={Pyramidal flow matching for efficient video generative modeling},
  author={Jin, Yang and Sun, Zhicheng and Li, Ningyuan and Xu, Kun and Jiang, Hao and Zhuang, Nan and Huang, Quzhe and Song, Yang and Mu, Yadong and Lin, Zhouchen},
  journal={arXiv preprint arXiv:2410.05954},
  year={2024}
}

@article{alonso2024diffusion,
  title={Diffusion for world modeling: Visual details matter in atari},
  author={Alonso, Eloi and Jelley, Adam and Micheli, Vincent and Kanervisto, Anssi and Storkey, Amos J and Pearce, Tim and Fleuret, Fran{\c{c}}ois},
  journal={Advances in Neural Information Processing Systems},
  volume={37},
  pages={58757--58791},
  year={2024}
}

@article{gao2024vid,
  title={Vid-gpt: Introducing gpt-style autoregressive generation in video diffusion models},
  author={Gao, Kaifeng and Shi, Jiaxin and Zhang, Hanwang and Wang, Chunping and Xiao, Jun},
  journal={arXiv preprint arXiv:2406.10981},
  year={2024}
}

@article{henschel2024streamingt2v,
  title={Streamingt2v: Consistent, dynamic, and extendable long video generation from text},
  author={Henschel, Roberto and Khachatryan, Levon and Poghosyan, Hayk and Hayrapetyan, Daniil and Tadevosyan, Vahram and Wang, Zhangyang and Navasardyan, Shant and Shi, Humphrey},
  journal={arXiv preprint arXiv:2403.14773},
  year={2024}
}

@article{wu2022nuwa,
  title={Nuwa-infinity: Autoregressive over autoregressive generation for infinite visual synthesis},
  author={Wu, Chenfei and Liang, Jian and Hu, Xiaowei and Gan, Zhe and Wang, Jianfeng and Wang, Lijuan and Liu, Zicheng and Fang, Yuejian and Duan, Nan},
  journal={arXiv preprint arXiv:2207.09814},
  year={2022}
}

@article{chen2024diffusion,
  title={Diffusion forcing: Next-token prediction meets full-sequence diffusion},
  author={Chen, Boyuan and Mart{\'\i} Mons{\'o}, Diego and Du, Yilun and Simchowitz, Max and Tedrake, Russ and Sitzmann, Vincent},
  journal={Advances in Neural Information Processing Systems},
  volume={37},
  pages={24081--24125},
  year={2024}
}

@article{song2025history,
  title={History-Guided Video Diffusion},
  author={Song, Kiwhan and Chen, Boyuan and Simchowitz, Max and Du, Yilun and Tedrake, Russ and Sitzmann, Vincent},
  journal={arXiv preprint arXiv:2502.06764},
  year={2025}
}

@inproceedings{zhang2025mega,
  title={Mega: Memory-efficient 4d gaussian splatting for dynamic scenes},
  author={Zhang, Xinjie and Liu, Zhening and Zhang, Yifan and Ge, Xingtong and He, Dailan and Xu, Tongda and Wang, Yan and Lin, Zehong and Yan, Shuicheng and Zhang, Jun},
  booktitle={Proceedings of the IEEE/CVF International Conference on Computer Vision},
  pages={27828--27838},
  year={2025}
}

@article{yin2024slow,
  title={From slow bidirectional to fast autoregressive video diffusion models},
  author={Yin, Tianwei and Zhang, Qiang and Zhang, Richard and Freeman, William T and Durand, Fredo and Shechtman, Eli and Huang, Xun},
  journal={arXiv preprint arXiv:2412.07772},
  volume={2},
  year={2024}
}

@article{gu2025long,
  title={Long-Context Autoregressive Video Modeling with Next-Frame Prediction},
  author={Gu, Yuchao and Mao, Weijia and Shou, Mike Zheng},
  journal={arXiv preprint arXiv:2503.19325},
  year={2025}
}

@article{xie2024progressive,
  title={Progressive autoregressive video diffusion models},
  author={Xie, Desai and Xu, Zhan and Hong, Yicong and Tan, Hao and Liu, Difan and Liu, Feng and Kaufman, Arie and Zhou, Yang},
  journal={arXiv preprint arXiv:2410.08151},
  year={2024}
}

@article{chen2025skyreels,
  title={SkyReels-V2: Infinite-length Film Generative Model},
  author={Chen, Guibin and Lin, Dixuan and Yang, Jiangping and Lin, Chunze and Zhu, Juncheng and Fan, Mingyuan and Zhang, Hao and Chen, Sheng and Chen, Zheng and Ma, Chengchen and others},
  journal={arXiv preprint arXiv:2504.13074},
  year={2025}
}

@misc{magi1,
      title={MAGI-1: Autoregressive Video Generation at Scale},
      author={Sand-AI},
      year={2025},
      url={https://static.magi.world/static/files/MAGI_1.pdf},
}

@article{guo2023animatediff,
  title={Animatediff: Animate your personalized text-to-image diffusion models without specific tuning},
  author={Guo, Yuwei and Yang, Ceyuan and Rao, Anyi and Liang, Zhengyang and Wang, Yaohui and Qiao, Yu and Agrawala, Maneesh and Lin, Dahua and Dai, Bo},
  journal={arXiv preprint arXiv:2307.04725},
  year={2023}
}

@article{liu2024dynamics,
  title={Dynamics-Aware Gaussian Splatting Streaming Towards Fast On-the-Fly 4D Reconstruction},
  author={Liu, Zhening and Hu, Yingdong and Zhang, Xinjie and Song, Rui and Shao, Jiawei and Lin, Zehong and Zhang, Jun},
  journal={arXiv preprint arXiv:2411.14847},
  year={2024}
}

@inproceedings{sitzmann2021lfns,
               author = {Sitzmann, Vincent
                         and Rezchikov, Semon
                         and Freeman, William T.
                         and Tenenbaum, Joshua B.
                         and Durand, Fredo},
               title = {Light Field Networks: Neural Scene Representations
                        with Single-Evaluation Rendering},
               booktitle = {Proc. NeurIPS},
               year={2021}
            }

@inproceedings{yang2024direct,
  title={Direct-a-video: Customized video generation with user-directed camera movement and object motion},
  author={Yang, Shiyuan and Hou, Liang and Huang, Haibin and Ma, Chongyang and Wan, Pengfei and Zhang, Di and Chen, Xiaodong and Liao, Jing},
  booktitle={ACM SIGGRAPH 2024 Conference Papers},
  pages={1--12},
  year={2024}
}

@article{feng2024i2vcontrol,
  title={I2vcontrol-camera: Precise video camera control with adjustable motion strength},
  author={Feng, Wanquan and Liu, Jiawei and Tu, Pengqi and Qi, Tianhao and Sun, Mingzhen and Ma, Tianxiang and Zhao, Songtao and Zhou, Siyu and He, Qian},
  journal={arXiv preprint arXiv:2411.06525},
  year={2024}
}

@article{he2024cameractrl,
  title={Cameractrl: Enabling camera control for text-to-video generation},
  author={He, Hao and Xu, Yinghao and Guo, Yuwei and Wetzstein, Gordon and Dai, Bo and Li, Hongsheng and Yang, Ceyuan},
  journal={arXiv preprint arXiv:2404.02101},
  year={2024}
}

@article{he2025cameractrl,
  title={Cameractrl ii: Dynamic scene exploration via camera-controlled video diffusion models},
  author={He, Hao and Yang, Ceyuan and Lin, Shanchuan and Xu, Yinghao and Wei, Meng and Gui, Liangke and Zhao, Qi and Wetzstein, Gordon and Jiang, Lu and Li, Hongsheng},
  journal={arXiv preprint arXiv:2503.10592},
  year={2025}
}

@misc{gen3,
  author       = {Runway},
  title        = {Introducing Gen-3 Alpha: A new frontier for video generation},
  year         = {2024},
  url          = {https://runwayml.com/research/introducing-gen-3-alpha},
  note         = {Accessed: 2025-02-24}
}

@article{yu2025trajectorycrafter,
  title={Trajectorycrafter: Redirecting camera trajectory for monocular videos via diffusion models},
  author={YU, Mark and Hu, Wenbo and Xing, Jinbo and Shan, Ying},
  journal={arXiv preprint arXiv:2503.05638},
  year={2025}
}

@article{gu2025das,
        title={Diffusion as Shader: 3D-aware Video Diffusion for Versatile Video Generation Control}, 
         author={Zekai Gu and Rui Yan and Jiahao Lu and Peng Li and Zhiyang Dou and Chenyang Si and Zhen Dong and Qifeng Liu and Cheng Lin and Ziwei Liu and Wenping Wang and Yuan Liu},
         year={2025},
         journal={arXiv preprint arXiv:2501.03847}
        }

@inproceedings{ren2025gen3c,
    title={GEN3C: 3D-Informed World-Consistent Video Generation with Precise Camera Control},
    author={Ren, Xuanchi and Shen, Tianchang and Huang, Jiahui and Ling, Huan and
        Lu, Yifan and Nimier-David, Merlin and Müller, Thomas and Keller, Alexander and
        Fidler, Sanja and Gao, Jun},
    booktitle={Proceedings of the IEEE/CVF Conference on Computer Vision and Pattern Recognition},
    year={2025}
}

@article{yang2025omnicam,
  title={OmniCam: Unified Multimodal Video Generation via Camera Control},
  author={Yang, Xiaoda and Xu, Jiayang and Luan, Kaixuan and Zhan, Xinyu and Qiu, Hongshun and Shi, Shijun and Li, Hao and Yang, Shuai and Zhang, Li and Yu, Checheng and others},
  journal={arXiv preprint arXiv:2504.02312},
  year={2025}
}

@article{bai2025recammaster,
  title={Recammaster: Camera-controlled generative rendering from a single video},
  author={Bai, Jianhong and Xia, Menghan and Fu, Xiao and Wang, Xintao and Mu, Lianrui and Cao, Jinwen and Liu, Zuozhu and Hu, Haoji and Bai, Xiang and Wan, Pengfei and others},
  journal={arXiv preprint arXiv:2503.11647},
  year={2025}
}

@article{luo2025camclonemaster,
  title={CamCloneMaster: Enabling Reference-based Camera Control for Video Generation},
  author={Luo, Yawen and Bai, Jianhong and Shi, Xiaoyu and Xia, Menghan and Wang, Xintao and Wan, Pengfei and Zhang, Di and Gai, Kun and Xue, Tianfan},
  journal={arXiv preprint arXiv:2506.03140},
  year={2025}
}

@inproceedings{yu2024wonderjourney,
  title={Wonderjourney: Going from anywhere to everywhere},
  author={Yu, Hong-Xing and Duan, Haoyi and Hur, Junhwa and Sargent, Kyle and Rubinstein, Michael and Freeman, William T and Cole, Forrester and Sun, Deqing and Snavely, Noah and Wu, Jiajun and others},
  booktitle={Proceedings of the IEEE/CVF Conference on Computer Vision and Pattern Recognition},
  pages={6658--6667},
  year={2024}
}

@inproceedings{
    engstler2024invisible,
    title={Invisible Stitch: Generating Smooth 3D Scenes with Depth Inpainting},
    author={Paul Engstler and Andrea Vedaldi and Iro Laina and Christian Rupprecht},
    year={2024},
    booktitle={Arxiv}
}

@article{yu2024wonderworld,
    title={WonderWorld: Interactive 3D Scene Generation from a Single Image},
    author={Hong-Xing Yu and Haoyi Duan and Charles Herrmann and William T. Freeman and Jiajun Wu},
    journal={arXiv:2406.09394},
    year={2024}
}

@article{xu2024easyanimate,
  title={Easyanimate: A high-performance long video generation method based on transformer architecture},
  author={Xu, Jiaqi and Zou, Xinyi and Huang, Kunzhe and Chen, Yunkuo and Liu, Bo and Cheng, MengLi and Shi, Xing and Huang, Jun},
  journal={arXiv preprint arXiv:2405.18991},
  year={2024}
}

@article{allegro2024,
  title={Allegro: Open the Black Box of Commercial-Level Video Generation Model},
  author={Yuan Zhou and Qiuyue Wang and Yuxuan Cai and Huan Yang},
  journal={arXiv preprint arXiv:2410.15458},
  year={2024}
}

@article{fan2025vchitect,
  title={Vchitect-2.0: Parallel Transformer for Scaling Up Video Diffusion Models},
  author={Fan, Weichen and Si, Chenyang and Song, Junhao and Yang, Zhenyu and He, Yinan and Zhuo, Long and Huang, Ziqi and Dong, Ziyue and He, Jingwen and Pan, Dongwei and others},
  journal={arXiv preprint arXiv:2501.08453},
  year={2025}
}

@inproceedings{wang2024dust3r,
  title={Dust3r: Geometric 3d vision made easy},
  author={Wang, Shuzhe and Leroy, Vincent and Cabon, Yohann and Chidlovskii, Boris and Revaud, Jerome},
  booktitle={Proceedings of the IEEE/CVF Conference on Computer Vision and Pattern Recognition},
  pages={20697--20709},
  year={2024}
}

@misc{mast3r_eccv24,
      title={Grounding Image Matching in 3D with MASt3R}, 
      author={Vincent Leroy and Yohann Cabon and Jerome Revaud},
      booktitle = {ECCV},
      year = {2024}
}

@inproceedings{cabon2025must3r,
  title={Must3r: Multi-view network for stereo 3d reconstruction},
  author={Cabon, Yohann and Stoffl, Lucas and Antsfeld, Leonid and Csurka, Gabriela and Chidlovskii, Boris and Revaud, Jerome and Leroy, Vincent},
  booktitle={Proceedings of the Computer Vision and Pattern Recognition Conference},
  pages={1050--1060},
  year={2025}
}

@InProceedings{Yang_2025_Fast3R,
    title={Fast3R: Towards 3D Reconstruction of 1000+ Images in One Forward Pass},
    author={Jianing Yang and Alexander Sax and Kevin J. Liang and Mikael Henaff and Hao Tang and Ang Cao and Joyce Chai and Franziska Meier and Matt Feiszli},
    booktitle={Proceedings of the IEEE/CVF Conference on Computer Vision and Pattern Recognition (CVPR)},
    month={June},
    year={2025},
}

@article{wang2025continuous,
  title={Continuous 3D Perception Model with Persistent State},
  author={Wang, Qianqian and Zhang, Yifei and Holynski, Aleksander and Efros, Alexei A and Kanazawa, Angjoo},
  journal={arXiv preprint arXiv:2501.12387},
  year={2025}
}

@inproceedings{wang2024vggsfm,
  title={VGGSfM: Visual Geometry Grounded Deep Structure From Motion},
  author={Wang, Jianyuan and Karaev, Nikita and Rupprecht, Christian and Novotny, David},
  booktitle={Proceedings of the IEEE/CVF Conference on Computer Vision and Pattern Recognition},
  pages={21686--21697},
  year={2024}
}

@inproceedings{wang2025vggt,
  title={VGGT: Visual Geometry Grounded Transformer},
  author={Wang, Jianyuan and Chen, Minghao and Karaev, Nikita and Vedaldi, Andrea and Rupprecht, Christian and Novotny, David},
  booktitle={Proceedings of the IEEE/CVF Conference on Computer Vision and Pattern Recognition},
  year={2025}
}

@misc{keetha2025mapanything,
  title={{MapAnything}: Universal Feed-Forward Metric {3D} Reconstruction},
  author={Nikhil Keetha and Norman M\"{u}ller and Johannes Sch\"{o}nberger and Lorenzo Porzi and Yuchen Zhang and Tobias Fischer and Arno Knapitsch and Duncan Zauss and Ethan Weber and Nelson Antunes and Jonathon Luiten and Manuel Lopez-Antequera and Samuel Rota Bul\`{o} and Christian Richardt and Deva Ramanan and Sebastian Scherer and Peter Kontschieder},
  note={arXiv preprint arXiv:2509.13414},
  year={2025}
}

@article{valevski2024diffusion,
  title={Diffusion models are real-time game engines},
  author={Valevski, Dani and Leviathan, Yaniv and Arar, Moab and Fruchter, Shlomi},
  journal={arXiv preprint arXiv:2408.14837},
  year={2024}
}

@article{che2024gamegen,
  title={Gamegen-x: Interactive open-world game video generation},
  author={Che, Haoxuan and He, Xuanhua and Liu, Quande and Jin, Cheng and Chen, Hao},
  journal={arXiv preprint arXiv:2411.00769},
  year={2024}
}

@article{zhang2025matrixgame,
  title     = {Matrix-Game: Interactive World Foundation Model},
  author    = {Yifan Zhang and Chunli Peng and Boyang Wang and Puyi Wang and Qingcheng Zhu and Zedong Gao and Eric Li and Yang Liu and Yahui Zhou},
  journal   = {arXiv},
  year      = {2025}
}

@article{yang2024playable,
  title={Playable Game Generation},
  author={Yang, Mingyu and Li, Junyou and Fang, Zhongbin and Chen, Sheng and Yu, Yangbin and Fu, Qiang and Yang, Wei and Ye, Deheng},
  journal={arXiv preprint arXiv:2412.00887},
  year={2024}
}

@article{xiao2025worldmem,
  title={WORLDMEM: Long-term Consistent World Simulation with Memory},
  author={Xiao, Zeqi and Lan, Yushi and Zhou, Yifan and Ouyang, Wenqi and Yang, Shuai and Zeng, Yanhong and Pan, Xingang},
  journal={arXiv preprint arXiv:2504.12369},
  year={2025}
}

@article{oasis2024,
  title={Oasis: A universe in a transformer},
  author={Decart, Etched and McIntyre, Quinn and Campbell, Spruce and Chen, Xinlei and Wachen, Robert},
  journal={URL: https://oasis-model. github. io},
  year={2024}
}

@inproceedings{bruce2024genie,
  title={Genie: Generative interactive environments},
  author={Bruce, Jake and Dennis, Michael D and Edwards, Ashley and Parker-Holder, Jack and Shi, Yuge and Hughes, Edward and Lai, Matthew and Mavalankar, Aditi and Steigerwald, Richie and Apps, Chris and others},
  booktitle={Forty-first International Conference on Machine Learning},
  year={2024}
}

@article{parkerholder2024genie2,
  title         = {Genie 2: A Large-Scale Foundation World Model},
  author        = {Jack Parker-Holder and Philip Ball and Jake Bruce and Vibhavari Dasagi and Kristian Holsheimer and Christos Kaplanis and Alexandre Moufarek and Guy Scully and Jeremy Shar and Jimmy Shi and Stephen Spencer and Jessica Yung and Michael Dennis and Sultan Kenjeyev and Shangbang Long and Vlad Mnih and Harris Chan and Maxime Gazeau and Bonnie Li and Fabio Pardo and Luyu Wang and Lei Zhang and Frederic Besse and Tim Harley and Anna Mitenkova and Jane Wang and Jeff Clune and Demis Hassabis and Raia Hadsell and Adrian Bolton and Satinder Singh and Tim Rockt{\"a}schel},
  year          = {2024},
  url           = {https://deepmind.google/discover/blog/genie-2-a-large-scale-foundation-world-model/}
}

@article{feng2024matrix,
  title={The matrix: Infinite-horizon world generation with real-time moving control},
  author={Feng, Ruili and Zhang, Han and Yang, Zhantao and Xiao, Jie and Shu, Zhilei and Liu, Zhiheng and Zheng, Andy and Huang, Yukun and Liu, Yu and Zhang, Hongyang},
  journal={arXiv preprint arXiv:2412.03568},
  year={2024}
}

@article{yu2025gamefactory,
  title={GameFactory: Creating New Games with Generative Interactive Videos},
  author={Yu, Jiwen and Qin, Yiran and Wang, Xintao and Wan, Pengfei and Zhang, Di and Liu, Xihui},
  journal={arXiv preprint arXiv:2501.08325},
  year={2025}
}

@article{Brohan2023RT2VM,
  title={RT-2: Vision-Language-Action Models Transfer Web Knowledge to Robotic Control},
  author={Anthony Brohan and Noah Brown and Justice Carbajal and Yevgen Chebotar and Krzysztof Choromanski and Tianli Ding and Danny Driess and Kumar Avinava Dubey and Chelsea Finn and Peter R. Florence and Chuyuan Fu and Montse Gonzalez Arenas and Keerthana Gopalakrishnan and Kehang Han and Karol Hausman and Alexander Herzog and Jasmine Hsu and Brian Ichter and Alex Irpan and Nikhil J. Joshi and Ryan C. Julian and Dmitry Kalashnikov and Yuheng Kuang and Isabel Leal and Sergey Levine and Henryk Michalewski and Igor Mordatch and Karl Pertsch and Kanishka Rao and Krista Reymann and Michael S. Ryoo and Grecia Salazar and Pannag R. Sanketi and Pierre Sermanet and Jaspiar Singh and Anikait Singh and Radu Soricut and Huong Tran and Vincent Vanhoucke and Quan Ho Vuong and Ayzaan Wahid and Stefan Welker and Paul Wohlhart and Ted Xiao and Tianhe Yu and Brianna Zitkovich},
  journal={ArXiv},
  year={2023},
  volume={abs/2307.15818},
  url={https://api.semanticscholar.org/CorpusID:260293142}
}

@article{Li2023VisionLanguageFM,
  title={Vision-Language Foundation Models as Effective Robot Imitators},
  author={Xinghang Li and Minghuan Liu and Hanbo Zhang and Cunjun Yu and Jie Xu and Hongtao Wu and Chi-Hou Cheang and Ya Jing and Weinan Zhang and Huaping Liu and Hang Li and Tao Kong},
  journal={ArXiv},
  year={2023},
  volume={abs/2311.01378},
  url={https://api.semanticscholar.org/CorpusID:264935429}
}

@article{Kim2024OpenVLAAO,
  title={OpenVLA: An Open-Source Vision-Language-Action Model},
  author={Moo Jin Kim and Karl Pertsch and Siddharth Karamcheti and Ted Xiao and Ashwin Balakrishna and Suraj Nair and Rafael Rafailov and Ethan Paul Foster and Grace Lam and Pannag R. Sanketi and Quan Vuong and Thomas Kollar and Benjamin Burchfiel and Russ Tedrake and Dorsa Sadigh and Sergey Levine and Percy Liang and Chelsea Finn},
  journal={ArXiv},
  year={2024},
  volume={abs/2406.09246},
  url={https://api.semanticscholar.org/CorpusID:270440391}
}

@article{Hancock2024RuntimeOI,
  title={Run-time Observation Interventions Make Vision-Language-Action Models More Visually Robust},
  author={Asher Hancock and Allen Z. Ren and Anirudha Majumdar},
  journal={2025 IEEE International Conference on Robotics and Automation (ICRA)},
  year={2024},
  pages={9499-9506},
  url={https://api.semanticscholar.org/CorpusID:273098082}
}

@article{Zhen20243DVLAA3,
  title={3D-VLA: A 3D Vision-Language-Action Generative World Model},
  author={Haoyu Zhen and Xiaowen Qiu and Peihao Chen and Jincheng Yang and Xin Yan and Yilun Du and Yining Hong and Chuang Gan},
  journal={ArXiv},
  year={2024},
  volume={abs/2403.09631},
  url={https://api.semanticscholar.org/CorpusID:268385444}
}

@article{Li2025PointVLAIT,
  title={PointVLA: Injecting the 3D World into Vision-Language-Action Models},
  author={Chengmeng Li and Junjie Wen and Yan Peng and Yaxin Peng and Feifei Feng and Yichen Zhu},
  journal={ArXiv},
  year={2025},
  volume={abs/2503.07511},
  url={https://api.semanticscholar.org/CorpusID:276928085}
}

@article{Team2024OctoAO,
  title={Octo: An Open-Source Generalist Robot Policy},
  author={Octo Model Team and Dibya Ghosh and Homer Rich Walke and Karl Pertsch and Kevin Black and Oier Mees and Sudeep Dasari and Joey Hejna and Tobias Kreiman and Charles Xu and Jianlan Luo and You Liang Tan and Pannag R. Sanketi and Quan Vuong and Ted Xiao and Dorsa Sadigh and Chelsea Finn and Sergey Levine},
  journal={ArXiv},
  year={2024},
  volume={abs/2405.12213},
  url={https://api.semanticscholar.org/CorpusID:266379116}
}

@inproceedings{Wen2024DiffusionVLAGA,
  title={Diffusion-VLA: Generalizable and Interpretable Robot Foundation Model via Self-Generated Reasoning},
  author={Junjie Wen and Minjie Zhu and Yichen Zhu and Zhibin Tang and Jinming Li and Zhongyi Zhou and Chengmeng Li and Xiaoyu Liu and Yaxin Peng and Chaomin Shen and Feifei Feng},
  year={2024},
  url={https://api.semanticscholar.org/CorpusID:274465094}
}

@article{bai2023qwen,
  title={Qwen technical report},
  author={Bai, Jinze and Bai, Shuai and Chu, Yunfei and Cui, Zeyu and Dang, Kai and Deng, Xiaodong and Fan, Yang and Ge, Wenbin and Han, Yu and Huang, Fei and others},
  journal={arXiv preprint arXiv:2309.16609},
  year={2023}
}

@article{yang2025qwen3,
  title={Qwen3 technical report},
  author={Yang, An and Li, Anfeng and Yang, Baosong and Zhang, Beichen and Hui, Binyuan and Zheng, Bo and Yu, Bowen and Gao, Chang and Huang, Chengen and Lv, Chenxu and others},
  journal={arXiv preprint arXiv:2505.09388},
  year={2025}
}

@article{liu2024deepseek,
  title={Deepseek-v3 technical report},
  author={Liu, Aixin and Feng, Bei and Xue, Bing and Wang, Bingxuan and Wu, Bochao and Lu, Chengda and Zhao, Chenggang and Deng, Chengqi and Zhang, Chenyu and Ruan, Chong and others},
  journal={arXiv preprint arXiv:2412.19437},
  year={2024}
}

@article{team2025kimi,
  title={Kimi k2: Open agentic intelligence},
  author={Team, Kimi and Bai, Yifan and Bao, Yiping and Chen, Guanduo and Chen, Jiahao and Chen, Ningxin and Chen, Ruijue and Chen, Yanru and Chen, Yuankun and Chen, Yutian and others},
  journal={arXiv preprint arXiv:2507.20534},
  year={2025}
}

@article{kerbl20233d,
  title={3D Gaussian splatting for real-time radiance field rendering.},
  author={Kerbl, Bernhard and Kopanas, Georgios and Leimk{\"u}hler, Thomas and Drettakis, George},
  journal={ACM Trans. Graph.},
  volume={42},
  number={4},
  pages={139--1},
  year={2023}
}

@article{yang2025kwai,
  title={Kwai keye-vl 1.5 technical report},
  author={Yang, Biao and Wen, Bin and Ding, Boyang and Liu, Changyi and Chu, Chenglong and Song, Chengru and Rao, Chongling and Yi, Chuan and Li, Da and Zang, Dunju and others},
  journal={arXiv preprint arXiv:2509.01563},
  year={2025}
}

@inproceedings{liang2025referdino,
    title={ReferDINO: Referring Video Object Segmentation with Visual Grounding Foundations},
    author={Liang, Tianming and Lin, Kun-Yu and Tan, Chaolei and Zhang, Jianguo and Zheng, Wei-Shi and Hu, Jian-Fang},
    booktitle={Proceedings of the IEEE/CVF International Conference on Computer Vision},
    year={2025}
}

@article{lipman2022flow,
  title={Flow matching for generative modeling},
  author={Lipman, Yaron and Chen, Ricky TQ and Ben-Hamu, Heli and Nickel, Maximilian and Le, Matt},
  journal={arXiv preprint arXiv:2210.02747},
  year={2022}
}

@misc{zhang2023adding,
  title={Adding Conditional Control to Text-to-Image Diffusion Models}, 
  author={Lvmin Zhang and Anyi Rao and Maneesh Agrawala},
  booktitle={IEEE International Conference on Computer Vision (ICCV)},
  year={2023},
}

@article{zhou2018stereo,
  title={Stereo magnification: Learning view synthesis using multiplane images},
  author={Zhou, Tinghui and Tucker, Richard and Flynn, John and Fyffe, Graham and Snavely, Noah},
  journal={arXiv preprint arXiv:1805.09817},
  year={2018}
}

@article{wang2025spatialvid,
  title={Spatialvid: A large-scale video dataset with spatial annotations},
  author={Wang, Jiahao and Yuan, Yufeng and Zheng, Rujie and Lin, Youtian and Gao, Jian and Chen, Lin-Zhuo and Bao, Yajie and Zhang, Yi and Zeng, Chang and Zhou, Yanxi and others},
  journal={arXiv preprint arXiv:2509.09676},
  year={2025}
}

@article{duan2025worldscore,
  title={Worldscore: A unified evaluation benchmark for world generation},
  author={Duan, Haoyi and Yu, Hong-Xing and Chen, Sirui and Fei-Fei, Li and Wu, Jiajun},
  journal={arXiv preprint arXiv:2504.00983},
  year={2025}
}

@inproceedings{edstedt2024roma,
title={{RoMa: Robust Dense Feature Matching}},
author={Edstedt, Johan and Sun, Qiyu and Bökman, Georg and Wadenbäck, Mårten and Felsberg, Michael},
booktitle={IEEE Conference on Computer Vision and Pattern Recognition},
year={2024}
}

@article{ravi2024sam2,
  title={SAM 2: Segment Anything in Images and Videos},
  author={Ravi, Nikhila and Gabeur, Valentin and Hu, Yuan-Ting and Hu, Ronghang and Ryali, Chaitanya and Ma, Tengyu and Khedr, Haitham and R{\"a}dle, Roman and Rolland, Chloe and Gustafson, Laura and Mintun, Eric and Pan, Junting and Alwala, Kalyan Vasudev and Carion, Nicolas and Wu, Chao-Yuan and Girshick, Ross and Doll{\'a}r, Piotr and Feichtenhofer, Christoph},
  journal={arXiv preprint arXiv:2408.00714},
  url={https://arxiv.org/abs/2408.00714},
  year={2024}
}
}

\end{document}